\documentclass[10pt,journal,compsoc]{IEEEtran}

\usepackage{pifont}
\usepackage[american]{babel}
\usepackage{url}            %
\usepackage{bm}
\usepackage{microtype}      %
\usepackage{graphicx}
\usepackage{multirow}
\usepackage{amsmath}
\usepackage{amssymb}
\usepackage{algorithm,algorithmic}
\usepackage{diagbox}
\graphicspath{{figure/}}
\usepackage[utf8x]{inputenc}
\usepackage{textcomp}
\usepackage{hyperref}
\usepackage{color}

\usepackage[monochrome]{xcolor}

\ifCLASSOPTIONcompsoc

  \usepackage[nocompress]{cite}
\else
  \usepackage{cite}
\fi

\begin{document}
\title{Adversarial Learning of Structure-Aware Fully Convolutional Networks for Landmark Localization}
\author{Yu Chen$^1$, Chunhua Shen$^2$,
Hao Chen$^2$,
Xiu-Shen Wei$^3$, Lingqiao Liu$^2$ and Jian Yang$^1$

\IEEEcompsocitemizethanks{\IEEEcompsocthanksitem
$^1$Key Lab of Intelligent Perception and Systems for High-Dimensional Information of Ministry of Education,
and Jiangsu Key Lab of Image and Video Understanding for Social Security,  Nanjing University of Science and Technology;
$^2$School of Computer Science, The University of Adelaide, Australia; and Australian Centre for Robotic Vision;
$^3$Department of Computer Science and Technology, Nanjing University, China.
}
\IEEEcompsocitemizethanks{\IEEEcompsocthanksitem
This work was  done when Y. Chen and X-S. Wei were visiting The University of Adelaide.
Correspondence should be addressed to C. Shen and J. Yang.
}
}

\markboth{CHEN et al.: ADVERSARIAL POSENET}%
{Manuscript}

\IEEEtitleabstractindextext{%

\begin{abstract}
\leftskip=0pt \rightskip=0pt plus 0cm
Landmark/pose estimation in single monocular images has received much effort in computer vision due to its important applications. It remains a challenging task when input images come with severe occlusions caused by, e.g., adverse camera views. Under such
circumstances, biologically implausible pose predictions may be produced. In contrast, human vision is able to predict poses by exploiting
geometric constraints of landmark point inter-connectivity. To address the problem, by incorporating priors about the structure of pose
components, we propose a novel structure-aware fully convolutional network to implicitly take such priors into account during training of
the deep network. Explicit learning of such constraints is typically challenging. Instead, inspired by how human identifies implausible
poses, we design discriminators to distinguish the real poses from the fake ones (such as biologically implausible ones). If the pose
generator \textit{G} generates results that the discriminator fails to distinguish from real ones, the network successfully learns the priors.
Training of the network follows the strategy of conditional Generative Adversarial Networks (GANs). The effectiveness of the proposed
network is evaluated on three pose-related tasks: 2D human pose estimation, 2D facial landmark estimation and 3D human
pose estimation. The proposed approach significantly outperforms several state-of-the-art methods and almost always generates plausible
pose predictions, demonstrating the usefulness of implicit learning of structures using GANs.

\end{abstract}

}

\maketitle
\IEEEdisplaynontitleabstractindextext
\IEEEpeerreviewmaketitle

\tableofcontents
\clearpage

\vspace{0.1cm}


\section{Introduction}\label{sec:introduction}

\IEEEPARstart{L}andmark localization, a.k.a, keypoint localization, pose estimation or alignment (we use these terms interchangeably in the sequel), is a key step in many vision tasks.
For example, face alignment, which is to locate the positions of a set of predefined facial landmarks from a single monocular facial image, plays an important role for facial augmented reality and face recognition.
Human pose estimation locates the positions of a few human body joints, which is critically important in understanding the actions and emotions of people in images and videos.
Keypoint prediction from monocular images is a challenging task due to factors such as high flexibility of facial/body limbs deformation, self and outer occlusion, various camera angles, etc.
In this work, we consider the problem of human pose estimation and facial landmark detection in the same framework with minimum modification as essentially they both are image-to-point regression problems. We achieved state-of-the-art on both tasks at the submission of this manuscript.

Recently, significant improvements have been achieved on 2D pose estimation by using Deep Convolutional Neural Networks (DCNNs)~\cite{jourabloo2017pose,trigeorgis2016mnemonic,kowalski2017deep,conf/nips/TompsonJLB14,conf/cvpr/TompsonGJLB15,conf/cvpr/ToshevS14,conf/cvpr/ChuOLW16,conf/cvpr/WeiRKS16,conf/eccv/NewellYD16,conf/eccv/BulatT16}.
These approaches mainly follow the strategy of regressing heatmaps or landmark coordinates of each pose part using DCNNs.
These regression models have shown great ability in learning better feature representations.
However, for pose components with heavy occlusions or background clutters that appear similar to body parts, DCNNs may have difficulty in regressing accurate poses.

Human vision is capable of learning the shape structures from abundant observations.
Even under extreme occlusions, one can infer the potential poses and exclude the implausible ones.
It is, however, very challenging to incorporate the priors about shape structures into DCNNs, because, as pointed out in~\cite{conf/nips/TompsonJLB14}, the low-level mechanics of DCNNs is typically difficult to interpret, and DCNNs are most capable of learning features.

\begin{figure*}[!t]
\centering
\includegraphics[width=1.99\columnwidth]{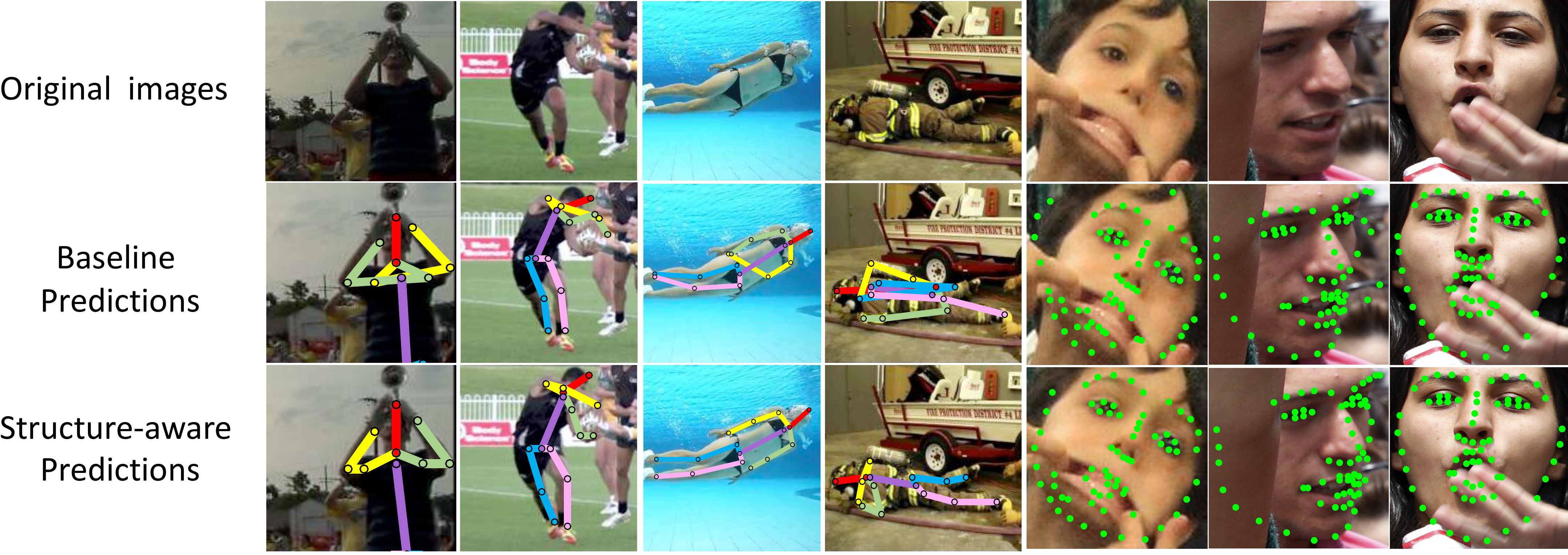}
\caption{\textbf{Motivation}. We show the importance of strongly enforcing priors about the pose structure during training of DCNNs for pose estimation. Learning without using such priors generates inaccurate results.}
\label{fig:The-blue-shape}
\end{figure*}

As a consequence, an unreasonable pose may be produced by conventional DCNNs.
As shown in Fig.~\ref{fig:The-blue-shape}, in challenging test cases with heavy occlusions, standard DCNNs tend to perform poorly.
To tackle this problem, priors about the structure of the body joints should be taken into account.
The key to this problem is to learn the {\em real} body joints distribution from a large amount of training data.
However, explicit learning of such a distribution is not trivial.

To address this problem, we attempt to learn the distribution of the human body structures {\em implicitly}.
Similar to the human vision, we suppose that we have a ``discriminator'' which can tell whether the predicted pose is geometrically plausible.
If the DCNN regressor is able to ``deceive'' the ``discriminator'' that its predictions are all reasonable, the network would have successfully learned the priors of the human body structure.

Inspired by the recent success in Generative Adversarial Networks (GAN)~\cite{journals/corr/RadfordMC15,journals/corr/ZhaoML16,conf/nips/SalimansGZCRCC16,conf/nips/GoodfellowPMXWOCB14,conf/nips/DentonCSF15}, we propose to design the ``discriminator'' as the discriminator network in GAN while the regression network functions as the generative network.
Training the generator in an adversarial manner against the discriminator precisely meets our intention.

For both 2D human pose estimation and facial landmark localization, a baseline stacked bottom-up, top-down networks \textit{G} is designed to generate the pose heatmaps.
Based on the pose heatmaps, the pose discriminator (\textit{P}) is used to tell whether the pose configuration is plausible.
The generator is asked to ``fool'' the discriminators by training \textit{G} and \textit{P} in the generative adversarial manner.
Thus, the human body structure is implied in the \textit{P} net by guiding \textit{G} to the direction that is close to ground-truth heatmaps and satisfies joint-connectivity constraints of the human body.
The learned \textit{G} net is expected to be more robust to occlusions and cluttered backgrounds where the precise description for different body parts is required.

What is more, the function of the discriminator is not limited to heatmap regression based 2D pose estimation.
For tasks concerning structured outputs (\emph{e.g.}, 2D to 3D human pose transformation), we can easily extend our method by using the adversarial discriminator on a baseline method to learn the 3D structure distributions for generating more plausible 3D pose prediction, as we show in our experiments.

The main contributions of this work are thus as follows:
\begin{itemize}
\leftskip=0pt \rightskip=0pt plus 0cm
\item To our knowledge, we are the first to use Generative Adversarial Networks (GANs) to exploit the constrained pose distribution for improving pose estimation. We also design a stacked multi-task network for predicting both the pose heatmaps and the occlusion heatmaps to achieve improved results for 2D human pose estimation.

\item We design a novel network framework for pose estimation which takes the geometric constraints of keypoints connectivity into consideration. By incorporating the priors of the human body, prediction mistakes caused by occlusions and cluttered backgrounds are considerably reduced. Even when the network fails, the outputs of the network appear more like ``human'' predictions instead of ``machine'' predictions.

\item We evaluate our method on public 2D human pose estimation datasets, 2D facial landmark estimation datasets and 3D human pose estimation datasets. Our approach achieved state-of-the-art performance at the submission of this manuscript, and is able to consistently produce more plausible pose predictions compared to baseline methods.
\end{itemize}

Furthermore, concurrently with recent work of \cite{bulat2017far}, we may be one of the first to directly use DCNNs to regress heatmaps for facial landmark estimation.
Due to the help of the structure-aware network structure, the traditional complex cascaded procedure is avoided.

\begin{figure*}[!t]
\centering
\includegraphics[width=0.85\textwidth]{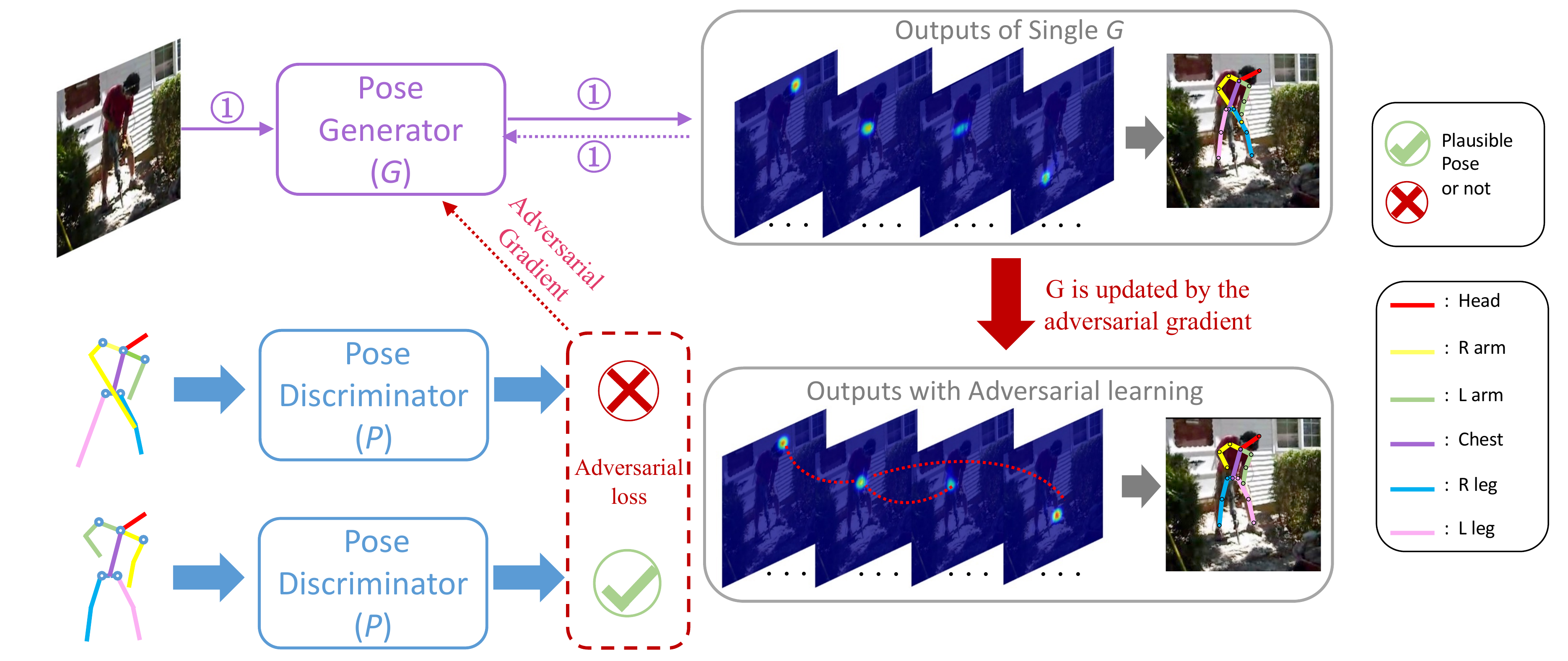}
\caption{Overview of the proposed Structure-aware Convolutional Network for human pose estimation. The sub-network in purple is the stacked multi-task network (\emph{G}) for pose generation. The networks in blue (\emph{P}) is used to discriminate whether the generated pose is ``real'' (reasonable as a body shape). The loss of \emph{G} has two parts: Mean Squared Error (MSE) of heatmaps (dashed line in purple) and Binary Cross Entropy (BCE) adversarial loss from \emph{P} (dashed line in red). Standalone training of \emph{G} produces results in the top-right. \emph{G} and \emph{P} produce results at the bottom-right.
}
\label{fig:Structure-of-the}
\end{figure*}

\section{Related Work} \label{sec:Related-Work}
The task of human pose estimation can be divided into multi-person and single-person.
Multi-person pose estimation involves both human detection and pose estimation.
The difficulty lies in accurate detection of individuals with different poses, overlapping or occlusions.
While in single-person human pose estimation, the rough positions of the person can be easily obtained.
The main challenge in single-person pose estimation is pose variation caused by body motion, etc.
As our method focuses on the positive influence of adversarial learning by exploiting the structure of pose estimation, we only consider single-person pose estimations and multi-person pose estimations with given person-detection results in this work.
In terms of vision tasks, our method is mostly related to 2D and 3D human pose estimation and 2D facial landmark
estimation issues.
In terms of mechanism of deep learning models, our method is mostly related to the Generative
Adversarial Networks.

\textbf{2D Human Pose Estimation.}
Traditional 2D single human pose estimation methods often follow the framework of tree structured graphical model~\cite{journals/ijcv/EichnerMZF12,journals/ijcv/BuehlerEHZ11,conf/eccv/SappTT10,conf/cvpr/YangR11,conf/iccv/PishchulinAGS13,conf/cvpr/SappT13}.
With the introduction of ``DeepPose'' by Toshev \emph{et al}.~\cite{conf/cvpr/ToshevS14}, deep network based methods become more popular in this area.
This work is closely related to the methods generating pose heatmaps from images~\cite{conf/cvpr/YangOLW16,conf/eccv/NewellYD16,conf/cvpr/TompsonGJLB15,conf/cvpr/WeiRKS16,conf/cvpr/ChuOLW16,conf/cvpr/PishchulinITAAG16,conf/eccv/InsafutdinovPAA16,conf/nips/TompsonJLB14}.
For example, Tompson \emph{et al}.~\cite{conf/nips/TompsonJLB14} used multi-resolution feature representations to generate heatmaps with joint-training of a Markov Random Field (MRF).
Tompson \emph{et al}.~\cite{conf/cvpr/TompsonGJLB15} used multiple branches of convolutional networks to fuse the features from an image pyramid, and used MRF for post-processing.
Later, Convolutional Pose Machine~\cite{conf/cvpr/WeiRKS16} incorporated
 the inference of the spatial correlations among body parts within convolutional networks.
The hourglass network~\cite{conf/eccv/NewellYD16} introduced a state-of-the-art architecture for bottom-up and top-down inference built upon residual blocks and skip connections.
Based on the hourglass structure, Chu \emph{et al}.~\cite{chu2017multi}
 used convolutional neural networks with a multi-context attention mechanism in an end-to-end framework.
The structure of our \textit{G} net for this task is also a fully convolutional network with ``conv-deconv'' architecture as in~\cite{conf/eccv/NewellYD16}.
However, our network is designed in a multi-task manner for improved performance.

Multi-person pose estimation methods mainly follow ``bottom-up" or ``top-down" architectures. The common ``top-down'' architecture \cite{gkioxari2014using,pishchulin2012articulated,sun2011articulated,huang2017coarse,he2017mask} is to use a person detector first and then employ single-person pose estimation methods. In comparison, ``bottom-up'' methods detect all joints first and then group them into different subjects. One of the most popular bottom-up methods is ~\cite{cao2017realtime} which proposed Part Affinity Fields(PAFs) to model the connectivity between joints. It is the first real-time multi-person estimation method and achieves best performance on the MSCOCO-2016 keypoint challenge. More recently, some top-down methods ~\cite{fang2017rmpe,chen2017cascaded} employ strong object detectors ~\cite{ren2015faster} and carefully-designed single-person pose estimation networks, which significantly outperform previous methods.
As our method does  not    involve any detection algorithm, we also follow the top-down
 approach with given detections to evaluate the effectiveness of adversarial learning.

{\bf 3D Human Pose Estimation.}
Based on the 2D human pose predictions, inferring 3D joints is to match the spatial position of the depicted person from 2D to 3D.
This can be traced back to the early work by Lee \emph{et al}. ~\cite{lee1985determination}.
As the literature of this problem is vast with approaches in a variety of settings \cite{scott1999factors}, here
we only review recent works which are most relevant to ours using deep networks in the sequel.

The first category is to infer 3D body configurations by estimating body angles from images~\cite{zhou2016deep,bogo2016keep}.
These approaches avoid estimating 3D joint positions directly, which offer the advantage of constraining the pose in a human-like structure and having lower dimensionality.
Recently, some systems have explored the possibility of directly inferring 3D poses from images with end-to-end deep architectures \cite{tekin2016structured}.
Pavlakos \emph{et al}.~\cite{pavlakos2016coarse} introduced a deep convolutional neural network
based on the stacked hourglass architecture~\cite{conf/eccv/NewellYD16}, which maps 2D joint probability heatmaps to probability distributions in the 3D space.
Moreno-Noguer \cite{moreno20163d} represented 2D and 3D poses with \textit{N}$\times$\textit{N} distance matrices (DMs)
and regresses 2D DMs to 3D DMs.

The DM regression approach as well as the the volumetric approach of Pavlakos \emph{et al.}
assumes that  direct regression from 2D keypoints to 3D keypoints is difficult.
However, Martinez \emph{et al}. \cite{martinez2017simple} showed that a simple fully-connected network can perform very well in the direct regression.
As this network has a simple structure and achieves high performance, we use it as our baseline model.
We demonstrate that the idea of enforcing the adversarial training  on the baseline model also works well for this 2D-to-3D problem.

More recently, Yang \emph{et al.} \cite{yang20183d} proposed an adversarial learning method. Their  method is  also built upon \cite{martinez2017simple} and has used the same training pipeline as in our method. The difference is that their method has three inputs: pose heatmaps, depth-maps and geometric descriptor,
which are jointly sent to the discriminator for determining if a pose is plausible. Their method is specially designed for 3D human pose estimation.

{\bf 2D Face Landmark Estimation.}
Traditional regression based methods often follow a cascaded manner to update the landmark localization results in a coarse-to-fine fashion.
This strategy has been proven to be  effective for face alignment.
Early methods mainly use random forest regression as the regressors due to computational efficiency~\cite{dollar2010cascaded,cao2014face,dantone2012real,valstar2010facial}.
Burgos-Artizzu et al.~\cite{burgos2013robust} proposed  Robust Cascaded Pose Regression (RCPR) which improves robustness to outliers by detecting occlusions explicitly.
Different from previous learning process, Supervised Descent Method (SDM) \cite{xiong2013supervised} attempted to directly minimize the feature deviation between estimated and ground-truth landmarks which was finally induced into a simple linear regression problem with supervised descent direction.
To accelerate the speed of SDM and overcome the drawbacks of handcrafted features, Local Binary Features (LBF) \cite{ren2014face} are learned for linear regression by using the regression forest.
Project-Out Cascaded Regression ~\cite{tzimiropoulos2015project} was proposed by learning and employing a sequence of averaged Jacobians and descent directions in a subspace orthogonal to the facial appearance variation.

Recently, deep neural networks were also introduced for face alignment \cite{sun2013deep,zhang2014coarse,chen2015facial,chen2016face}.
These methods use deep networks to replace the traditional regressors but still follow the cascaded framework.
It is worth pointing out that the Mnemonic Descent Method (MDM) ~\cite{trigeorgis2016mnemonic} showed that end-to-end  training of a convolutional recurrent neural network architecture works well for face alignment.
The original cascaded steps are connected by recurrent connections and  handcrafted features are replaced by  convolutional features.
We take a further step by directly regressing the landmark heatmaps from the face image.
This approach of direct regressing was considered inefficient and unrealistic by most
previous methods in the literature, as face shape is complex.
However, we show that with the help of the adversarial learning, shape priors can be better captured, and good localization results are
achieved.

{\bf Generative Adversarial Network.}
Generative Adversarial Networks have been widely studied in the literature for discrete labels~\cite{journals/corr/MirzaO14}, text~\cite{reed2016generative} and also images.
The conditional models have tackled inpainting~\cite{conf/cvpr/PathakKDDE16}, image prediction from a normal map~\cite{conf/eccv/WangG16}, future frame prediction~\cite{journals/corr/MathieuCL15}, future state prediction~\cite{conf/eccv/ZhouB16}, product photo generation~\cite{conf/eccv/YooKPPK16}, and style transfer~\cite{conf/eccv/LifshitzFU16}.

Human pose estimation can be considered as a translation from a RGB image to a multi-channel heatmap.
The designed bottom-up and top-down \textit{G} net can well accomplish this translation.
Different from previous work, the goal of the discrimination network is not only to distinguish the ``fake'' from ``real'', but also to incorporate geometric constraints into the model.
Thus we have implemented different training strategies for fake samples from traditional GANs.
In the next section, we provide details.

\section{Adversarial Learning for Landmark Localization}\label{sec:GcNet-for-Human}

As depicted in Fig.~\ref{fig:Structure-of-the}, the adversarial training model consists of two parts, \emph{i.e.},
the pose generator network \textit{G} and the pose discriminator network \textit{P}.
Without discriminators, \textit{G} will be updated simply by backward propagation of itself ({cf.}, the lines with \textcircled{1} in Fig.~\ref{fig:Structure-of-the}). This is defined as the {\it baseline model} for all the tasks.
Thus, incorrect location pose estimations may be generated.
It is necessary to leverage the power of discriminators to correct these incorrect estimations.
Therefore, a discriminator network \emph{P} is introduced into the framework.

After updating \textit{G} by training with \textit{P} in the adversarial manner (cf. the red dashed lines), the pose priors are implicitly exploited.
{\it In practical training, the two parts of the loss are added together to optimize for \textit{G} at the same time.}

Next, we first introduce the structure of the baseline generative networks and the discriminator networks. Then, we show the adversarial training paradigm.

\subsection{Generative Network}\label{subsec:Multi-task-Generative-Network}

In this section, we present the generative network \emph{G} (baseline model) in our framework. For 2D human pose estimation and facial landmark localization, the networks are fully-convolutional which predict pose estimations from images in an end-to-end manner. For 3D human pose estimation, we use a fully-connected network for 2D-to-3D coordinate transformation based upon 2D predictions.

\begin{figure*}[!t]
\centering
\includegraphics[width=2\columnwidth]{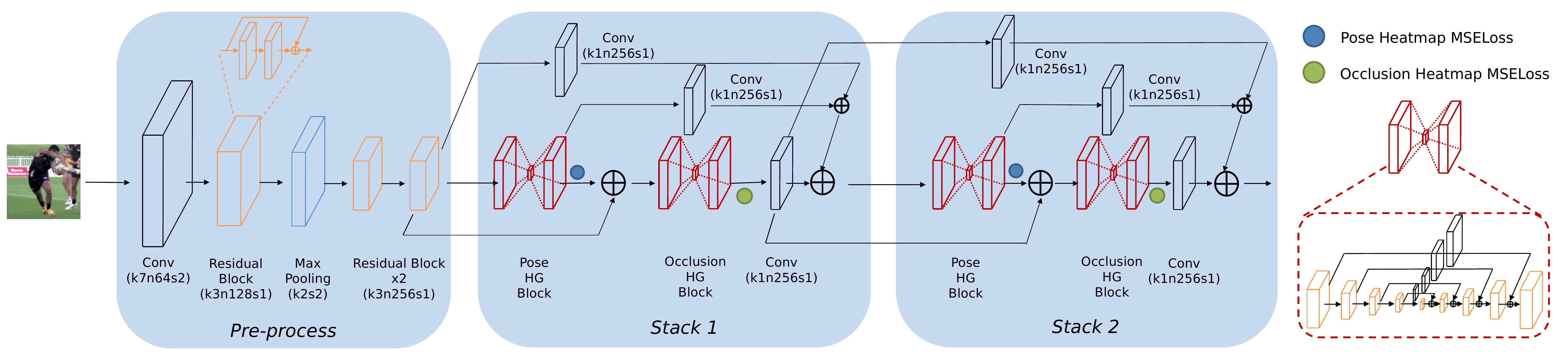}
\caption{Architecture of the multi-task generative network \textit{G}. Black, orange, blue and red rectangles indicate convolutional layers, residual blocks, max pooling layers and hourglass blocks respectively.  $\oplus$ indicates  addition of input features. Solid blue and green circles indicate pose and occlusion losses in the network. The brief architecture of the hourglass block is shown at the right. Stacking of the first and the second networks is displayed and more networks can be stacked with the same structure.}
\label{fig:Architecture-of-G}
\end{figure*}

\noindent \textbf{2D  Human  Pose Estimation.}
To solve the problem of human pose estimation, it can be very beneficial to employ local evidence
for identifying features for human joints.
Meanwhile, it clear to see that  a coherent understanding of the full body image must be in place to achieve good pose estimation.
In addition, as reported in~\cite{conf/cvpr/WeiRKS16}, large contextual regions are important for locating body parts.
Hence the contextual region of a neuron, which is its receptive field, should be large.
To achieve these goals, an ``encoder-decoder'' architecture is used.
Also, to capture information at each scale,  mirrored layers in the encoder and decoder ~\cite{conf/eccv/NewellYD16} are added, as shown in the bottom-right part of Fig.~\ref{fig:Architecture-of-G}.
Inspired by~\cite{conf/eccv/NewellYD16}, our network can also  be  stacked   to provide the network with the ability  to re-evaluate the previous estimates and features.
In each module of the \textit{G} net, a residual block~\cite{conf/cvpr/HeZRS16} is used for the convolution operator.

Besides, knowledge of whether a body part being occluded clearly offers important information for inferring the geometric information of a human pose.
Here, in order to effectively incorporate both pose estimation and occlusion predictions, we propose to tackle the problem with a multi-task generative network.
As shown in Fig.~\ref{fig:Architecture-of-G}, in each stacking module, poses and occlusions are jointly predicted. Then, both predictions are re-evaluated for the next stacking.

\def\x{ { \bm x } }

So the multi-task generative network is to learn a function $\mathcal{G}$ which attempts to project an image $\bm{x}$ to both the corresponding pose heatmaps $\bm{y}$ and occlusion heatmaps $\bm{z}$, \emph{i.e.}, $\mathcal{G}(\x) =  \{\hat{\bm{y}}, \hat{\bm{z}}\}$ where $\hat{\bm{y}}$ and $\hat{\bm{z}}$ are the predicted heatmaps.
Given the original image $\bm{x}$, a basic block of the stacked multi-task generator network can be expressed as follows:
\begin{equation}
\nonumber
\begin{cases}
\{\bm{Y}_{n},\bm{Z}_{n},\bm{X}\}=\mathcal{G}{}_{n}(\bm{Y}_{n-1},\bm{Z}_{n-1},\bm{X})  \quad{\rm if~~}n\geqslant2\\
\{\bm{Y}_{n},\bm{Z}_{n},\bm{X}\}=\mathcal{G}{}_{n}(\bm{X})\quad\quad\quad\quad\quad\quad~~ {\rm if~~} n=1
\end{cases}
\,,
\end{equation}
where $\bm{Y}_{n}$ and $\bm{Z}_{n}$ are the output activation tensors of the $n\textrm{-th}$ stacked generative network for pose estimations and occlusion predictions, respectively.
$\bm{X}$ is the image feature tensor, obtained after pre-processing on the original image through two residual blocks.
Suppose that there are $N$ times stacking of the basic block. The multi-task generative network can be formulated as:
\begin{equation}
\nonumber
\{\bm{Y}_{N},\bm{Z}_{N},\bm{X}\}=\mathcal{G}_{N}(\mathcal{G}_{N-1}(\cdots(\mathcal{G}_{1}(\bm{X}),\bm{Y}_{1},\bm{Z}_{1}))) \,.
\end{equation}
In each basic block, the final heatmap outputs $\hat{\bm{y}}_{n},\hat{\bm{z}}_{n}$ are obtained from $\bm{Y}_{n}$ and $\bm{Z}_{n}$ by two $1\times 1$ convolution layers with the step size of 1 and without padding. Specifically, the first convolution layer reduces the number of feature maps from the number of feature maps to the number of body parts. The second convolution layer acts as a linear classifier to obtain the final predicted heatmaps.

Therefore, given a training set $\{\bm{x}^{i},\bm{y}^{i},\bm{z}^{i}\}_{i=1}^{M}$ where $M$ is the number of training images, the loss function of our multi-task generative network is presented as:
\begin{equation}
\label{eq:multitaskG}
\mathcal{L}_{G}(\Theta)=\frac{1}{2MN}\sum_{n=1}^{N}\sum_{i=1}^{M}\left(\left\Vert \bm{y}^{i}-\hat{\bm{y}}_{n}^{i}\right\Vert ^{2}+\left\Vert \bm{z}^{i}-\hat{\bm{z}}_{n}^{i}\right\Vert ^{2}\right) \,,
\end{equation}
where $\Theta$ denotes the parameter set.

\noindent \textbf{2D  Facial Landmark Localization.}
In contrast to most previous methods which predicts facial landmark location as coordinates, we use the same heatmap regression approach as human pose estimation.
The variations of face shapes are clearly less complicated than human poses and most facial landmark databases do not
contain visibility annotations.
Therefore, we remove the occlusion heatmap regression part in Fig.~\ref{fig:Architecture-of-G} as the baseline for facial landmark localization.
Thus, the network becomes a stacked hourglass architecture which is the same as in ~\cite{conf/eccv/NewellYD16}.

\noindent \textbf{3D   Human   Pose Estimation.}
In this paper, 3D human poses are not predicted from scratch and are used as an extended validation of adversarial learning for 3D structure learning.
To be specific, we follow ~\cite{martinez2017simple} which deals 3D pose estimation as a 2-step procedure.
First, heatmap based 2D predictions are given using fully-convolutional networks.
Then 2D coordinates are extracted by extracting the locations of the maximum values in the heatmaps.
Finally the 2D-to-3D coordinate transformation is done by combinations of linear layers followed by batch normalization, dropout and ReLU activation functions.
To fully understand the network structure, readers may refer to ~\cite{martinez2017simple} for details.

\subsection{Discriminative Networks}\label{subsec:Pose-Discriminator}

To enable the training of the network to exploit priors about the body joints configurations, we design the pose discriminator \textit{P}.
The role of the discriminator \emph{P} is to distinguish the {\it fake} poses---those poses
do not satisfy the constraints of pose components---from the {\it real} poses.

\vspace{0.3em}
\noindent\textbf{2D   Pose   Discriminator.}
It is intuitive that we need local image regions to identify the body parts and the large image patches (or the whole image) to understand the relationships between body parts.
However, when some parts are seriously occluded, it can  be very difficult  to locate the body parts. Human can achieve that by  using prior knowledge and observing both the local image patches around the body parts and relationships among different body parts.
Inspired by this, both low-level and high-level information can be important to infer whether the predicted poses are biologically plausible.
In contrast to previous work, we use an encoder-decoder architecture to implement the discriminator \emph{P}.
Skip connections between parallel layers are used to incorporate both the local and global information.

\begin{figure}[!t]
\centering
\includegraphics[width=0.95\columnwidth]{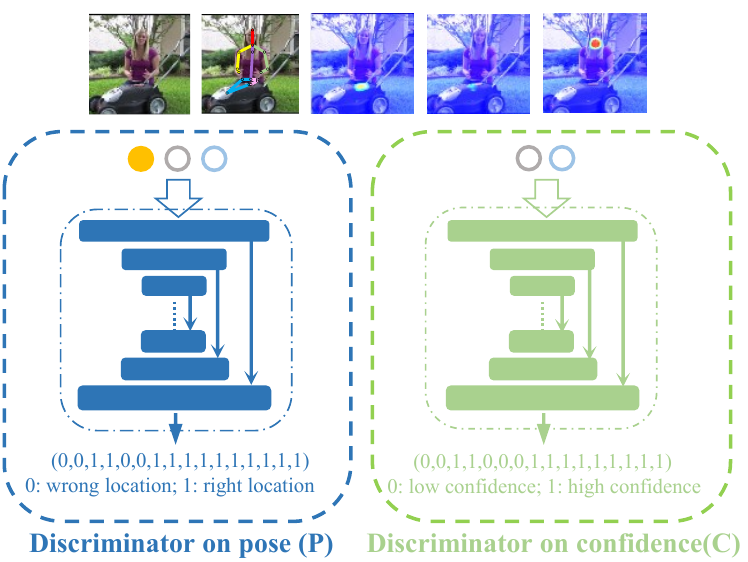}
\caption{Architectures of the 2D pose discriminator networks \textit{P} and \textit{C}. On the top we show the image for pose estimation, the image with estimated joints and heatmaps of right ankle, pelvis and neck (1st, 7th and 9th of all pose heatmaps respectively). The expected output for this sample is given at the bottom of the dashed box.}
\label{fig:Architectures-of-P}
\end{figure}

Additionally, even when the generative network fails to predict the correct pose locations for a particular image, the predicted pose may still be a plausible one for another human body shape.
Thus, simply using the pose and occlusion features may still face difficulty in training an accurate \textit{P}.
{\it Such inference should be made by taking the original image into consideration at the same time.}
When occlusion information can be provided, it is also helpful in inferring the pose rationality.
Thus we use the input RGB image with pose and occlusion heatmaps generated by the \textit{G} net as the inputs to \textit{P} for predicting whether a pose is reasonable for 2D human pose estimation.
The network structure of \textit{P} is shown in Fig.~\ref{fig:Architectures-of-P}.
To achieve this goal, GAN is designed in the conditional manner for \textit{P} in our framework.
As GANs learn a generative model of data, conditional GANs (cGANs) learn a conditional generative model~\cite{conf/nips/GoodfellowPMXWOCB14}.
The objective of a conditional adversarial \textit{P} network is expressed as follows:
\begin{align}
\label{eq:disP}
\mathcal{L}_{P}(G,P) &= \mathbb{E}[\textrm{log}P(\bm{y},\bm{z},\bm{x})]  +   \notag  \\
  &  \mathbb{E}[\textrm{log}(1-\vert P(G(\bm{x}),\bm{x})  -\bm{p}_{\rm  fake} \vert)],
\end{align}
where $\bm{p}_{\rm  fake}$ is the ground truth pose discriminator label. In traditional GAN,  $\bm{p}_{\rm  fake}$ is simply set to  0.
Detailed discussions of $\bm{p}_{\rm  fake}$
are presented in Section~\ref{subsec:Geometric-Constrained-Training}.

\begin{figure}[t!]
\centering
\includegraphics[width=1\columnwidth]{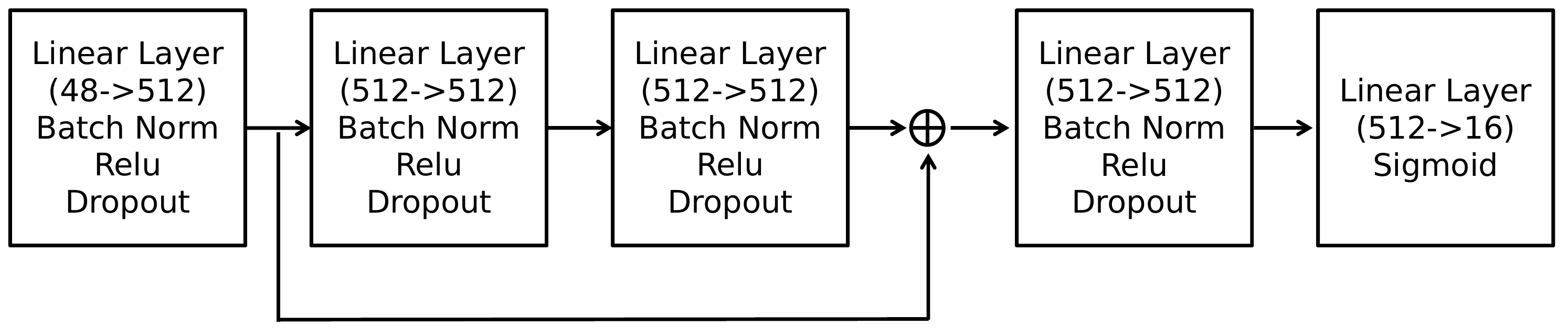}
\caption{Structure of the 3D pose discriminator when a 16-joints pose is to be discriminated. In each black rectangle, a few modules are combined.}
\label{fig:3d_Dis}
\end{figure}

\vspace{0.3em}
\noindent \textbf{3D   Pose  Discriminator.}
Different from 2D pose estimations, 3D poses are in the form of 3-dimensional coordinates.
In consistency with the simple structure in 2D-to-3D transformation, a five-layer fully connected network is used as the 3D Pose Discriminator as shown in Fig.~\ref{fig:3d_Dis}.
The output of this discriminator is the same as the one for 2D and also follows the same objective functions as in Eq.~\eqref{eq:disP}.

\vspace{0.3em}
\noindent \textbf{Auxiliary 2D Confidence Discriminator for 2D Human Pose.}
For the task of human pose estimation, some body parts are out of the boundary of the images and the corresponding joints are not required to be predicted.
So the confidence of the heatmap has an additional function to predict whether the joint is in the image.
On the other hand, by observing the differences between ground truth heatmaps and predicted heatmaps by previous methods, we find that the predicted ones are often not Gaussian centered because of occlusions and body overlapping.
Recalling the mechanism of human vision, even when the body parts are occluded, we can still confidently locate the body parts.
That is mainly because we already acquire the geometric prior of human body joints.
Motivated by this, we design a second  auxiliary discriminator, which is termed Confidence Discriminator (\emph{i.e.}, \textit{C}) to discriminate the high-confidence predictions from the low-confidence predictions.
The inputs for \textit{C} are the pose and occlusion heatmaps.
The objective of a traditional adversarial \textit{C} network can be expressed as:
\begin{align}
\label{eq:disC}
\mathcal{L}_{C}(G,C)&=\mathbb{E}[\textrm{log}C(\bm{y},\bm{z})]+  \notag  \\
& \mathbb{E}[\textrm{log}(1-\vert C(G(\bm{x})) -\bm{c}_{\rm  fake} \vert)] \,.
\end{align}
where $\bm{c}_{\rm  fake}$ is the ground truth confidence label. In traditional GAN,  $\bm{c}_{\rm  fake}$ is simply set as 0. The illustration of $\bm{c}_{\rm  fake}$ here will also be discussed in Section~\ref{subsec:Geometric-Constrained-Training}.

\subsection{Training of the Adversarial Networks}\label{subsec:Geometric-Constrained-Training}

In this section, we describe in detail how discriminators contribute to the accurate pose predictions with structure constraints.

First we show how to embed the geometric information of human bodies into the proposed \textit{P} network.
We observe that, when a part of human body is occluded, the prediction of the un-occluded parts are typically not affected.
This may be due to the DCNN's strong ability in learning local features.

However, in previous works on image translation using GANs, the discriminative network is learned with  all fake samples being labeled as negative samples.
When predicted heatmaps are sufficiently close  to ground-truths, considering it as a successful prediction makes sense. We also find the network to be difficult to converge by simply setting 0 or 1 as the ground truth label for a sample.
Based on these observations, we design a novel strategy for pose estimation. This leads to the difference with traditional GANs as in Eq.~\eqref{eq:disP} and Eq.~\eqref{eq:disC}.

\begin{algorithm}[t]
\caption{The training process of our method.}
\label{alg:The-training-process}
\small
\begin{algorithmic}[1]{
\REQUIRE {Training images:  $\bm{x}$, the corresponding ground-truth heatmaps \{\textit{$\bm{y}$,$\bm{z}$}\}};
\STATE {Forward \textit{P} by $\{\hat{\bm{p}}_{\rm fake}\}=\emph{P}(\bm{x},\emph{G}(\bm{x}))$, and optimize \textit{P} net by maximizing the second term in Eq.~\eqref{eq:disP}};
\STATE {Forward \textit{P} by $\{\hat{\bm{p}}_{\rm real}\}=\emph{P}(\bm{x},\bm{y},\bm{z})$, and optimize \textit{P} by maximizing the first term in Eq.~\eqref{eq:disP}};
\STATE {Forward \textit{C} by $\{\hat{\bm{c}}_{\rm fake}\}=\emph{C}(\emph{G}(\bm{x}))$, and optimize \textit{C} by maximizing the second term in Eq.~\eqref{eq:disC}};
\STATE {Forward \textit{C} by $\{\hat{\bm{c}}_{\rm real}\}=\emph{C}(\bm{y},\bm{z})$, and optimize \textit{C} by maximizing the first term in Eq.~\eqref{eq:disC}};
\STATE {Optimize \textit{G} by Eq.~\eqref{eq:final}};
\STATE {Go back to \textbf{Step 1} until convergence (one may check on the validation set)};
\RETURN {\emph{G}}.
}\end{algorithmic}
\end{algorithm}

The ground truth $\bm{p}_{\rm real}$ of a real sample is a $16\times 1$ (16 is the number of body parts) vector filled with $1$.
For the fake samples, if a predicted body part is far from the ground truth location, the pose is clearly implausible for the body configuration in this image.
Therefore, when training P, the ground truth $\bm{p}_{\rm fake}$ is:
\begin{equation}
\nonumber
\bm{p}_{\rm fake}^{i}=\begin{cases}
1 & {\rm if}\thinspace d_{i}<\delta\\
0 & {\rm if}\thinspace d_{i}\geqslant\delta
\end{cases}\,,
\end{equation}
where $\delta$ is the threshold parameter and $d_{i}$ is the normalized distance between the predicted and ground-truth location of the $i$-$\textrm{th}$ body part.
The range of the output values in \textit{P} is also $\left[0,1\right]$.
To deceive \textit{P}, \textit{G} will be trained to generate heatmaps
that satisfy the joints constraints of human bodies.

As mentioned in Section~\ref{subsec:Pose-Discriminator}, auxiliary confidence discriminator \textit{C} is required for 2D human pose estimation.
If \textit{G} generates low-confidence heatmaps, \textit{C} would  classify the result as ``fake''.
As \textit{G} is optimized to deceive \textit{C} that the fakes are being real, this process would help \textit{G} to generate high confidence heatmaps even with occlusions presented.
The outputs are the confidence scores $\bm{c}$ which in fact corresponds to whether the network is confident in locating body parts.

During training \textit{C}, the real heatmaps are labelled with  a $16\times 1$ vector $\bm{c}_{\rm real}$ filled with $1$. The confidence of the fake (predicted) heatmap should be high when it is close to ground truth and low otherwise, instead of being low for all predicted heatmaps as in traditional GANs.
Therefore the fake (predicted) heatmaps are labeled with a   $16\times 1$ vector $\bm{c}_{\rm fake}$ where the elements of $\bm{c}_{\rm fake}$ are the corresponding confidence scores.

\begin{equation}
\nonumber
\bm{c}_{\rm fake}^{i}=\begin{cases}
1 & {\rm if}\thinspace\left\Vert \bm{y}_{i}-\hat{\bm{y}_{i}}\right\Vert <\varepsilon\\
0 & {\rm if}\thinspace\left\Vert \bm{y}_{i}-\hat{\bm{y}_{i}}\right\Vert \geqslant\varepsilon
\end{cases}\,,
\end{equation}
where $\varepsilon$ is the threshold parameter, and $i$ is the $i$-th body part. The range of the output values in \textit{C} is $\left[0,1\right]$.

 Previous approaches to conditional GANs have found it beneficial to mix the GAN objective with a  traditional loss, such as $\ell_2$ distance~\cite{conf/cvpr/PathakKDDE16}. For our task, it is clear  that we also need to supervise \textit{G} in the training process with the ground truth poses. Thus, the discriminator still plays the original role, but the generator will not only fool the discriminator but also approximate the ground-truth output in an $\ell_2$ sense as in Eq.~\eqref{eq:disC}. Therefore, the final objective function is presented as follows.
\begin{equation}
\label{eq:final}
\arg \min_G \max_{P,C}  \mathcal{L}_{G}(\Theta)+\mathcal{\alpha L}_{C}(G,C)+\mathcal{\beta L}_{P}(G,P) \,.
\end{equation}
Here $\alpha=0$ \rm{if} $\bm{c}_{\rm fake}=\bm{c}_{\rm real}$, $\beta=0$ \rm{if} $\bm{p}_{\rm fake}=\bm{p}_{\rm real}$.
In experiments, in order to make the different components of the final objective function have the same scale, the hyper parameters $\alpha$ and $\beta$ are set to $1/220$ and $1/180$, respectively.
Algorithm~\ref{alg:The-training-process} demonstrates the whole training processing as the pseudo codes.
For training tasks without \textit{C}, we can simply set $\alpha$ to $0$.

\begin{figure*}[!t]
\centering
\includegraphics[width=0.4\textwidth]{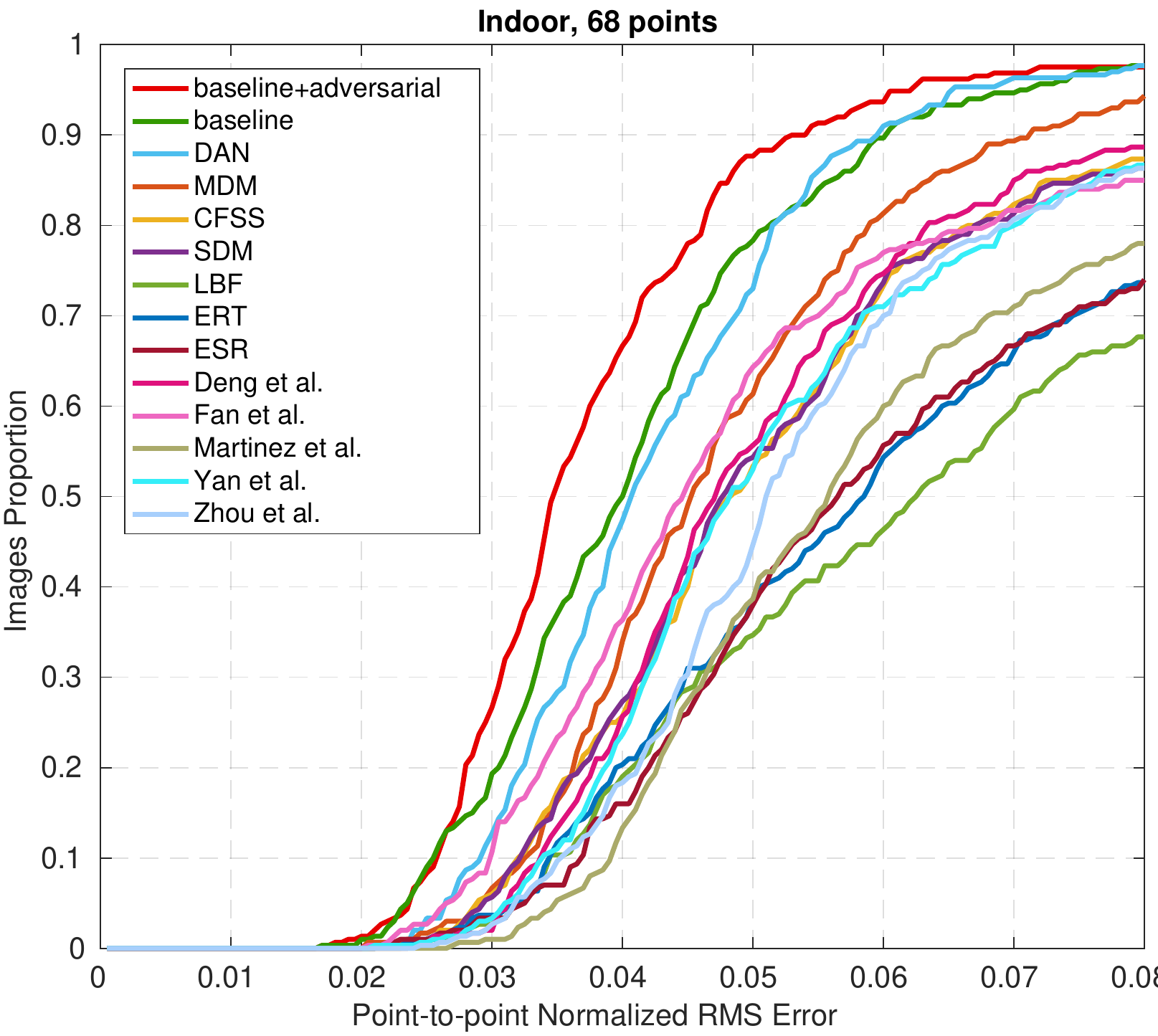}
\hspace{1em}
\includegraphics[width=0.39\textwidth]{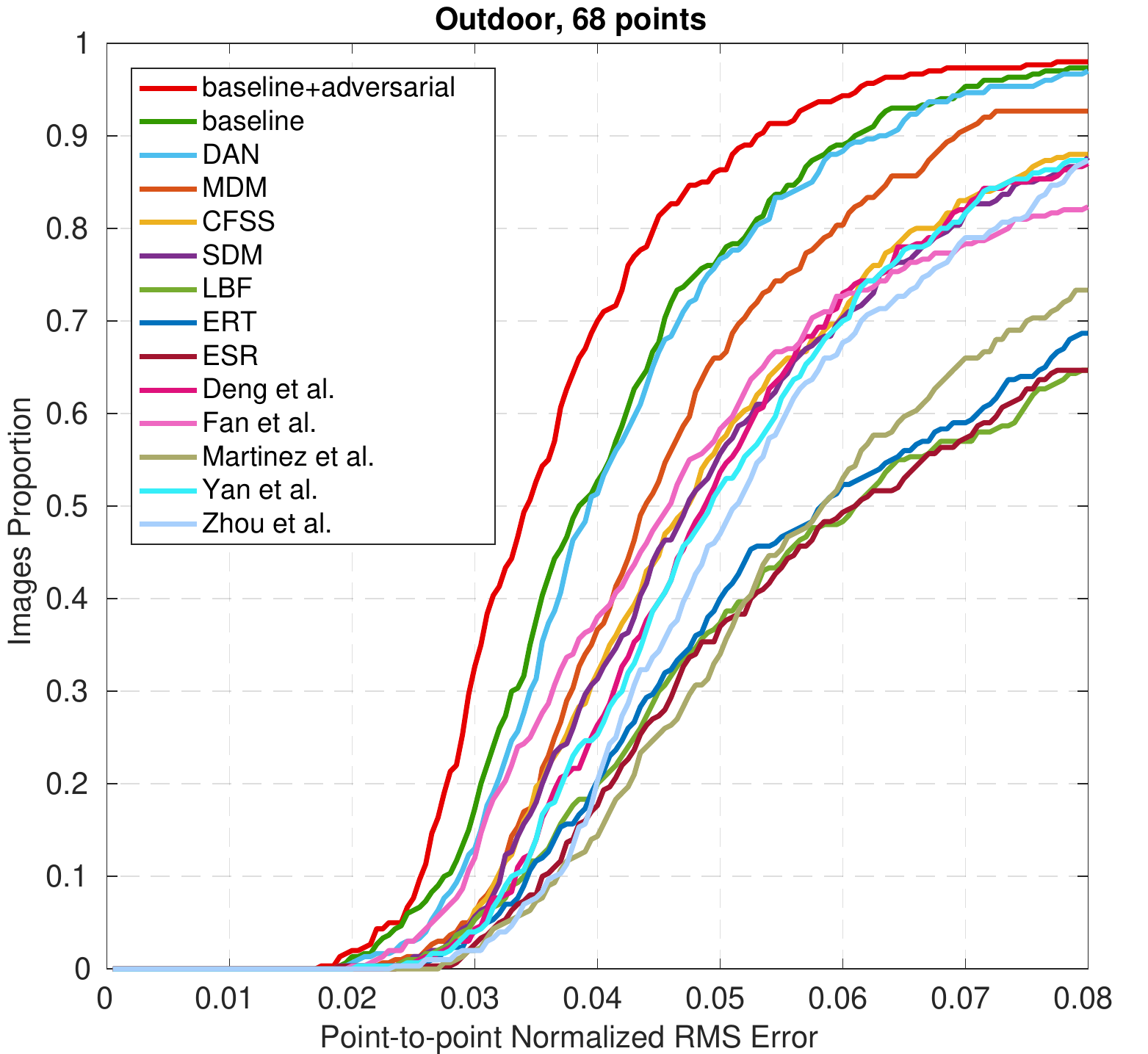}
\caption{Quantitative results on the test set of the 300W competition (indoor and outdoor) for 68-point prediction.The point-to-point error is normalized by the inter-ocular distance.}
\label{fig:CED}
\end{figure*}

\begin{figure}[!t]
\centering
\includegraphics[width=0.4\textwidth]{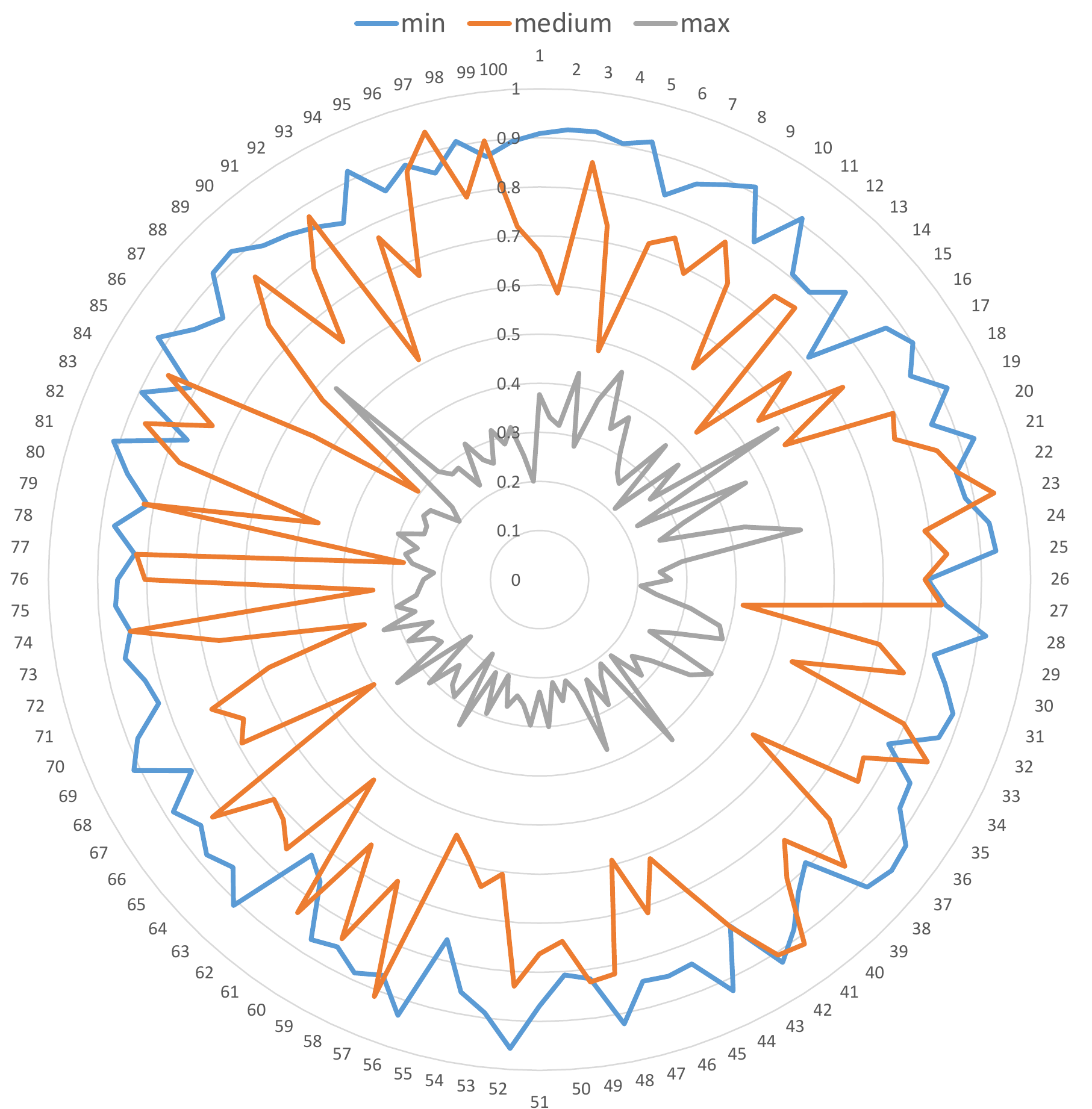}
\caption{Results of the discriminator network \textit{P} for the task of facial landmark estimation. The samples are sorted from the highest NRMSE error to the lowest one. The discriminator scores of the top 100 samples are marked with the gray line. The medium 100 samples are marked with the orange line. The lowest 100 samples are marked with the blue line.}
\label{fig:Fig_P}
\end{figure}

\section{Experiments}\label{sec:Experiments}

We evaluate the effectiveness of the proposed adversarial learning strategy on three structural tasks: 2D facial landmark detection, 2D single human pose estimation and 2D-to-3D human pose transformation.

\subsection{Facial Landmark Detection}\label{subsec:exp_face}
\noindent \textbf{Datasets.}
There are different strategies of annotating landmarks in the literature, such as 5 key points~\cite{sun2013deep}, 21 key points ~\cite{koestinger11a}, 29 key points~\cite{belhumeur2013localizing} and 68 key points~\cite{sagonas2013300}.
We follow the 68-points annotating as the main experimental setting as the level of difficulty of estimation increases with more landmarks.
The annotations are provided for LFPW~\cite{belhumeur2013localizing}, HELEN~\cite{le2012interactive}, AFW~\cite{zhu2012face} and IBUG~\cite{sagonas2013300} datasets.

The details of these datasets are as follows: ~\textit{(i)} 811 training images and 224 testing images in LFPW, ~\textit{(ii)} 2000 training images and 330 testing images in HELEN, ~\textit{(iii)} 337 images in AFW, ~\textit{(iv)} 135 images in IBUG.
These databases are used for training of our method.
As the official test set of 300W competition~\cite{sagonas2013300} was not released at first, the testing images in LFPW and HELEN is commonly refereed as the ~\textit{common test set of 300W competition~\cite{sagonas2013300}}, the images in IBUG is commonly refereed as the ~\textit{challenging test set of 300W}.
The common and challenging test sets together are referred as the ~\textit{full test set of 300W}.
After the later version of 300W competition, the official test set consisting of 300 indoor and 300 outdoor images was released, which was reported to have similar configuration as the IBUG dataset.
In our method, we follow the standard routine to use images in LFPW, HELEN, AFW and IBUG for training and 600 official test images for testing.
All annotations and bounding boxes are available at \url{https://ibug.doc.ic.ac.uk/resources/facial-point-annotations/}.

Furthermore, we conduct an ablation experiment on the AFLW dataset ~\cite{koestinger11a} with 21 landmarks since it contains more non-frontal faces.
We follow the experimental settings in ~\cite{lv2017deep,zhu2016unconstrained} where landmarks of two ears are not estimated.
As in ~\cite{zhu2016unconstrained}, the dataset is split into two sets: AFLW-Full and AFLW-Frontal. The full set contains 20,000 training faces and 4,386 testing faces. The frontal set uses the same training set but only uses 1,165 frontal faces for evaluation.

\noindent \textbf{Experimental Settings.} According to the estimated bounding boxes of faces, we use the center location and the diagonal distance of the bounding box to crop the face images into similar scales at the resolution of 256$\times$256 pixels. To make the network robust to different face initialization, we follow the popular routine ~\cite{zhang2014coarse,trigeorgis2016mnemonic} to augment samples by (0.75-1.25) scaling and $\pm$25$^\circ$ in-plane rotations generated from a uniform distribution. To reduce computation consumption, the network starts with a 7$\times$7 convolutional layer with stride 2 to downsize the resolution  from 256$\times$256 to 128$\times$128. Then the proposed network is connected to the 128 feature maps.  The networks is stacked four times in this task. For implementation, we train all our 2D pose models with the Torch7 toolbox~\cite{collobert2011torch7}.

\begin{table}[t!]
	\caption{Comparisons of mean error, AUC and failure rate (at a threshold of 0.08 of the normalized error) on the 300W test dataset.}
        \label{tab:Comparisons-of-AUC}
	\small
	\centering
	\setlength{\tabcolsep}{8.0pt}
	\renewcommand{\arraystretch}{1.1}
	\begin{tabular}{ r||c|c|c }
		\hline
			{Methods} & \emph{Mean error (\%)} & \emph{AUC} & \emph{Failure (\%)}  \\
		\hline
		\hline
		ESR \cite{cao2014face}  & 8.47  & 26.09   &  30.50  \\
		ERT\cite{kazemi2014one}   & 8.41  & 27.01   & 28.83   \\
		LBF ~\cite{ren2014face}\footnotemark[1]  	& 8.57   & 25.27   & 33.67    \\
		Yan \emph{et al.}\cite{yan2013learn} & -  & 34.79   & 12.67   \\
		Face++ \cite{zhou2013extensive}  & -  &  32.23  &  13.00  \\
		SDM \cite{xiong2013supervised}   & 5.83    & 36.27   & 13.00    \\
		CFAN \cite{zhang2014coarse}  & 5.78   & 34.78  & 14.00     \\
		CFSS \cite{zhu2015face}   & 5.74  & 36.58  &  12.33    \\
		MDM \cite{trigeorgis2016mnemonic}   & 4.78   & 45.32  & 6.80     \\
		DAN \cite{kowalski2017deep}   & 4.30   & 47.00   & 2.67     \\
		\hline
		\hline
		Baseline & \textbf{4.25}& \textbf{50.06}& \textbf{2.67}\\
		Ours & \textbf{3.96}& \textbf{53.64}& \textbf{2.50}\\
		\hline
	\end{tabular}
	\leftline{
	\footnotemark[1]{The implementation uses the fast version of LBF.}}
\end{table}

\begin{table}[!t]
 \caption{Results on the AFLW facial landmark detection test set.} \label{tab:AFLW}
 \small
 \centering
\setlength{\tabcolsep}{2.5pt}
 \renewcommand\arraystretch{1}
 \begin{tabular}{ r |c|c|c|c|c|c|c }
  \hline
  \multirow{1}{*}{Methods} & \multicolumn{1}{c}{SDM} & \multicolumn{1}{c}{ERT}  & \multicolumn{1}{c}{LBF} & \multicolumn{1}{c}{CFSS} & {SAN~\cite{dong2018style}} &\multicolumn{1}{c}{Baseline}  & \multicolumn{1}{c}{Ours}\\
  \hline
AFLW-Full  & \multicolumn{1}{c}{4.05} &  \multicolumn{1}{c}{4.35}  &  \multicolumn{1}{c}{4.25}  &  \multicolumn{1}{c}{3.92} &  {1.91} &  \multicolumn{1}{c}{1.81} &  \multicolumn{1}{c}{\textbf{1.39}}\\
AFLW-Front &  \multicolumn{1}{c}{2.94}  &  \multicolumn{1}{c}{2.75}  &  \multicolumn{1}{c}{2.74}  &  \multicolumn{1}{c}{2.68}&   {1.85} & \multicolumn{1}{c}{1.67}  &  \multicolumn{1}{c}{\textbf{1.32}}\\
  \hline
 \end{tabular}
\end{table}

\subsubsection{Quantitative Results}\label{subsec:face_result}

We follow the same protocol of reporting errors as the 300w competition, where the average point-to-point Euclidean error normalized by the inter-ocular distance (measured as the Euclidean distance between the outer corners of the eyes) is used as the error measure.

First, we report our results in the form of CED curves in Fig. \ref{fig:CED} which is consistent with~\cite{sagonas2013300}. Our method is compared against a few  state-of-the-art methods, including
Deep Alignment Network (DAN)~\cite{kowalski2017deep}, Mnemonic Descent Method~\cite{trigeorgis2016mnemonic}, Coarse-to-Fine Shape Searching (CFSS)~\cite{zhu2015face}, Coarse-to-Fine Auto-encoder Networks (CFAN)~\cite{zhang2014coarse}, Local Binary Features (LBF)~\cite{ren2014face}, Explicit Regression Trees (ERT)~\cite{kazemi2014one}, Supervised Descent Method (SDM)~\cite{xiong2013supervised}, Explicit Shape Regression (ESR) \cite{cao2014face}, Deng \emph{et al.}~\cite{deng2016m}, Fan \emph{et al.}~\cite{fan2016approaching}, Martinez \emph{et al.}~\cite{martinez20162}, Uricar\emph{et al.}~\cite{martinez20162}, Face++ ~\cite{zhou2013extensive} and Yan \emph{et  al.}~\cite{yan2013learn}.

Results of the last six methods as listed are quoted from the 300W competition website.
SDM is implemented by our-self  using the dense-SIFT feature provided by the author of the original paper.
For other methods, publicly available implementations are used for testing.
The results  demonstrate that our method outperforms compared face alignments methods in every error metrics.
The adversarial learning strategy clearly
improves the performance compared to the baseline model.

It should be noted that,  although our method avoids the coarse-to-fine approaching strategy, we perform much better in fine estimation.
In particular, compared to MDM, which uses the CNN features with a recurrent process to replace the original cascaded modules, our method uses stacked modules instead and achieves better results.
Compared to the insistence of cascaded strategy before, this sets a new point of view that CNN is capable of end-to-end learning such a complex and accurate regression function for face alignment,
if the network's capacity is sufficient, and more importantly,
we exploit  supervision  appropriately for training.

We have calculated a few more metrics from the CED curve to provide insights
into the performance of our method, such as mean error, area-under-the-curve (AUC) and the failure rate (at a threshold of 0.08 of the normalized error) of each method.
Only the top three performing methods of the last competition are shown in the table, as in Table ~\ref{tab:Comparisons-of-AUC}. It can be shown that, although our method improves
marginally
in error rate when the threshold is set at 0.08, our method greatly reduces the mean error and improves the AUC performance
significantly.

For AFLW, we use the face size  to normalize the mean error as the evaluation metric. The performance is reported in Table~\ref{tab:AFLW}.
Our method is compared against some of the methods in Table ~\ref{tab:Comparisons-of-AUC} and the
Style Aggregated Network (SAN) ~\cite{dong2018style}.
The results of other methods on AFLW are quoted from
\cite{dong2018style}. We follow the same train/valid split and show better performance than other methods.
Besides, the adversarial learning strategy again clearly improves the performance of our baseline model especially
for \textit{non-frontal faces}.

\subsubsection{Qualitative Comparisons}\label{subsec:face_comparison}
\begin{figure*}[!t]
\centering
\includegraphics[width=1\textwidth]{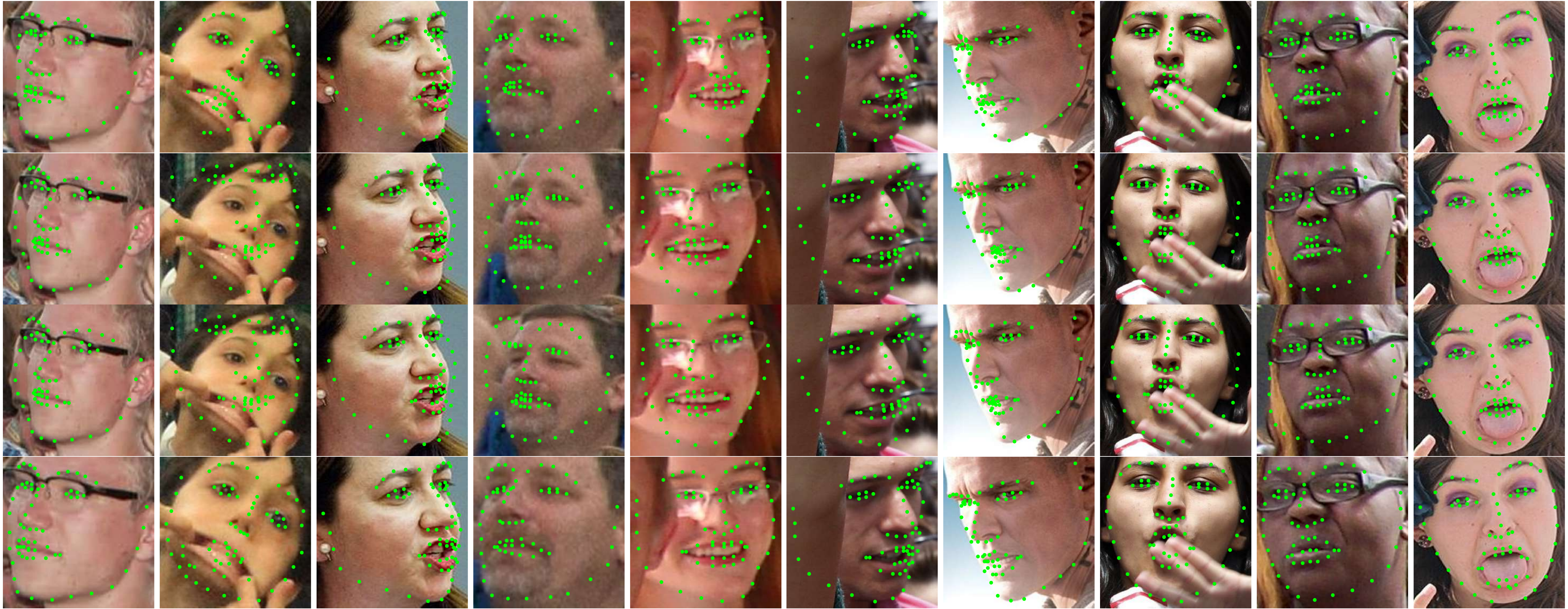}
\caption{Samples on the 300W test set. The four rows are results of MDM \cite{trigeorgis2016mnemonic}, CFSS
\cite{zhu2015face}, SDM\cite{xiong2013supervised} and our method respectively. After estimation by each method, the coordinates are projected to the original image. Then the images are cropped to make sure that all the estimated landmarks are within the displayed image, which results in different scales of the displayed images. }
\label{fig:Samples on 300W}
\end{figure*}

To intuitively show the improvement of our method over previous methods, we show samples with large errors using
previous methods in Fig.~\ref{fig:Samples on 300W}.
It can be easily observed that
{\it our method estimates more reasonable face shapes under extreme poses and occlusions.
}
For example, in the second column, CFSS and SDM fail to locate most of the landmarks which produce a set of disordered points.
Although MDM succeeds  to locate the landmarks without occlusions, it fails
in the part of occluded mouth and surrounding face contour.
Especially for the face contour, the landmarks are  estimated    without explicitly enforcing
 shape constraints.
On the other side, our method succeeds  in locating the landmarks accurately and maintains plausible  face shapes.

A comparison to the baseline model is also given with faces of large occlusions in Fig.~\ref{fig:The-blue-shape}.
We observe that the baseline model can accurately locate most landmarks under occlusions.
However, as geometric information is lacking, a few landmarks are clearly implausible.
Our adversarial learning strategy can fix this problem.

To further show the usefulness of the discriminator network, we display the result scores in Fig.~\ref{fig:Fig_P}.
As the generator network has been trained to successfully ``deceive''  the discriminator,
the estimates of the final network are fairly accurate, which corresponds to a low failure rate on the 300W test set.
Discrimination results for these estimations are mostly be extremely high, which would not help in terms of
observing the usefulness of \textit{P}.

Hence, we use a non-fully-converged intermediate generator network for evaluation.
As the test set of 300W only contain 600 images, to show the results more clearly, we use another divided database: 300VW ~\cite{chrysos2015offline,shen2015first,tzimiropoulos2015project} for evaluation.
We uniformly sample 4,397 images from the original video images.
The intermediate generator network is used to estimate the landmark predictions.
Then the predictions are sent into the final discriminator network to get the discrimination scores.
In Fig.~\ref{fig:Samples on 300W}, we clearly observe  that the low scores well correspond to the predictions with large errors, while the high scores correspond to the ones with small errors.
This indicates the discrimination capability of our  designed discriminator.
As long as the generator successfully ``deceive''
this discriminator, the landmark estimations become more accurate.

\def\etal{{\it et al.}}

\begin{table}[t!]
	\caption{Comparisons of PCK@0.2 performance on the LSP dataset.}
        \label{tab:Comparisons-of-PCK}
	\small
	\centering
	\setlength{\tabcolsep}{3.0pt}
	\scalebox{0.875}
	{
	\begin{tabular}{ r ||c|c|c|c|c|c|c||c  }
		\hline
			{Methods} & \emph{Head} & \emph{Sho.} & \emph{Elb.} & \emph{Wri.} & \emph{Hip} & \emph{Knee} & \emph{Ank.} & \textbf{Mean} \\
		\hline
		\hline
		B.\&A.'17 \cite{Belagiannis2016} & 95.2   & 89.0    & 81.5  &  77.0  &  83.7   &  87.0   & 82.8  &  85.2  \\
		Lifshitz'16 \cite{conf/eccv/LifshitzFU16}  & 96.8  &  89.0  &  82.7  &  79.1  &  90.9  &  86.0   & 82.5   & 86.7  \\
		Pishchulin'13~\cite{conf/iccv/PishchulinAGS13}   & 97.0    & 91.0   & 83.8   & 78.1   & 91.0  &  86.7  &  82.0   & 87.1  \\
		Insafutdinov'16 \cite{conf/eccv/InsafutdinovPAA16}   & 97.4   & 92.7   & 87.5  &  84.4  &  91.5  &  89.9   & 87.2  &  90.1  \\
		Pishchulin'16~\cite{conf/cvpr/PishchulinITAAG16}   & 97.8   & 92.5   & 87.0   & 83.9  &  91.5   & 89.9  &  87.2  &  90.1  \\
		Wei'16~\cite{conf/cvpr/WeiRKS16}   & 97.8   & 92.5  &  87.0  &  83.9  &  91.5   & 90.8   & 89.9   & 90.5  \\
		B.\&T.'16 \cite{conf/eccv/BulatT16}   & 97.2   & 92.1   & 88.1   & 85.2   & 92.2   & 91.4  &  88.7   & 90.7  \\
			Chou'17~{\cite{chou2017self}\footnotemark[1]}  & {{98.2}}& {94.9}& {92.2}& {89.5}& {94.2}& {95.1}&
				{94.1}&{94.0}\\
		\hline
		\hline
		Ours & \textbf{98.5}& \textbf{94.0}& \textbf{89.8}& \textbf{87.5}& \textbf{93.9}& \textbf{94.1}& \textbf{93.0}&\textbf{93.1}\\
		\hline
	\end{tabular}
	}
	\vspace{1em}
\leftline{
\footnotemark[1]{\footnotesize{\textcolor{black}{Published after the submission of our conference version.}}}}
\end{table}

\subsection{2D Human Pose Estimation}\label{subsec:exp_pose}
\noindent \textbf{Datasets.} We evaluate the proposed method on three widely used benchmarks on pose estimation, \emph{i.e.}, extended Leeds Sports Poses (LSP)~\cite{Johnson10}, MPII Human Pose~\cite{conf/cvpr/AndrilukaPGS14} for single-person human pose estimation and MSCOCO  Keypoints dataset ~\cite{lin2014microsoft} for multi-person human pose estimation.

The LSP dataset consists of 11,000 training images and 1,000 testing images from sports activities.
The MPII dataset consists of around 25,000 images with 40,000 annotated samples (about 28,000 for training, 11,000 for testing).
The figures are annotated with 16 landmarks on the whole body with various challenging directions to the camera. On MPII, we train our model on a subset of training images while evaluating on the official test set and a held-out validation set about 3,000 samples~\cite{conf/cvpr/TompsonGJLB15,conf/eccv/NewellYD16}.

For the MSCOCO keypoint estimation dataset, 17 body joints are annotated with around 15,000 subjects.
As our method doesn't involve any human detection module, we base our experiments on fixed detection results by an object detector algorithm based on FPN \cite{lin2017feature} provided by \cite{chen2017cascaded}.
Our method is trained on the MSCOCO 2017 training set and MPII multi-person dataset.
The performance is tested on the MSCOCO 2017 test-dev set.
All datasets provide the visibility of body parts, which are used as the supervision occlusion signal in our method.

\begin{figure}[t!]
\centering
\includegraphics[width=0.95\columnwidth]{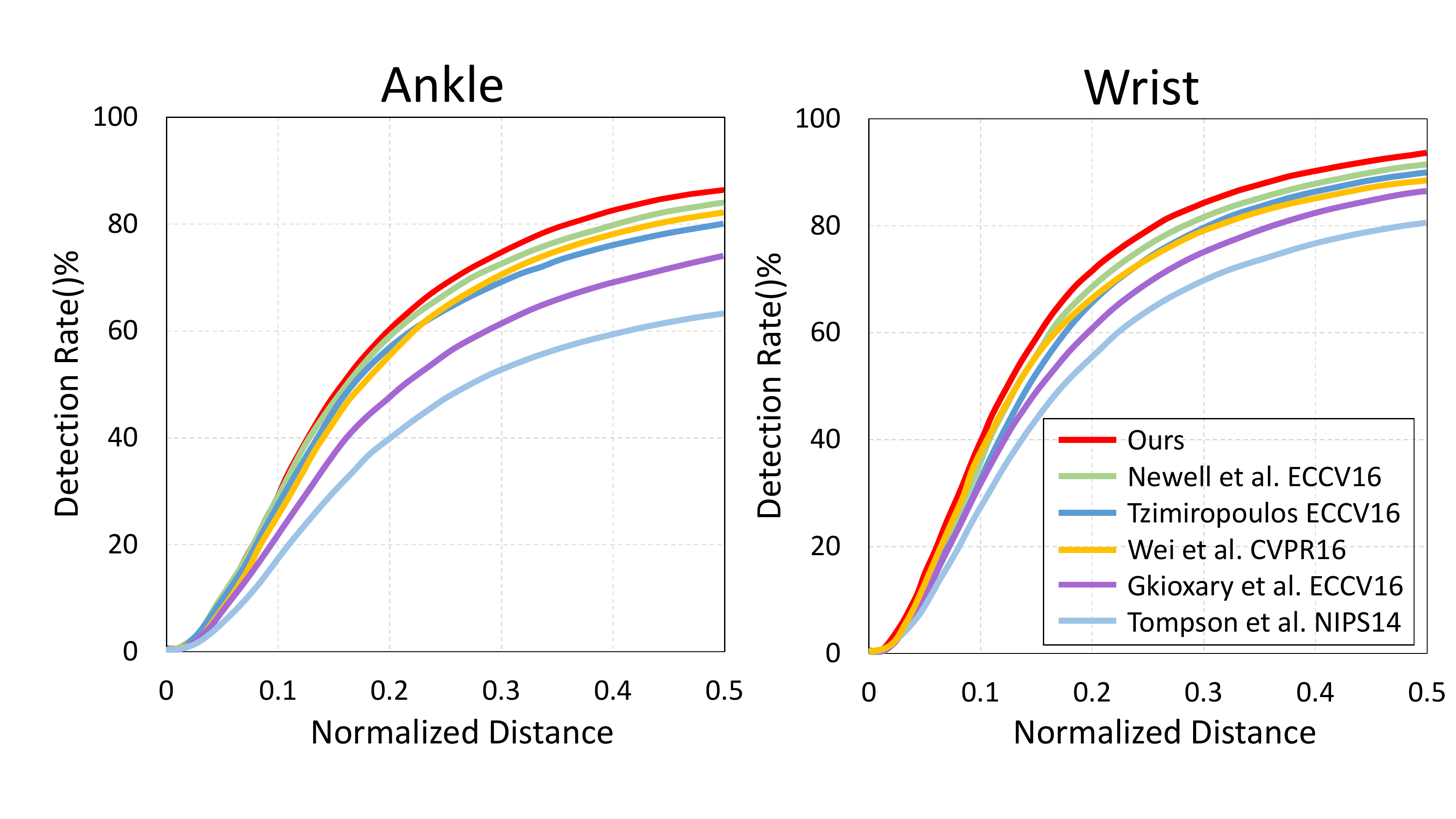}
\caption{PCKh comparison on MPII validation set.}
\label{fig:PCKh-comparison-on}
\end{figure}

\begin{table}[t!]
 \caption{Results on MPII Human Pose (PCKh@0.5).}
 \label{tab:Results-on-MPII}
 \small
 \centering
 \setlength{\tabcolsep}{2.0pt}
 \begin{tabular}{ r ||c|c|c|c|c|c|c||c }
  \hline
   {Methods} & \emph{Head} & \emph{Sho.} & \emph{Elb.} & \emph{Wri.} & \emph{Hip} & \emph{Knee} & \emph{Ank.} & \textbf{Mean} \\
  \hline
  \hline
  Tompson'14 \cite{conf/nips/TompsonJLB14} & 95.8 &  90.3  & 80.5  & 74.3  & 77.6  & 69.7  & 62.8 & 79.6\\
  Carreira \etal~\cite{conf/cvpr/CarreiraAFM16}  & 95.7 &  91.7 &  81.7 &  72.4 &  82.8  & 73.2  & 66.4 & 81.3\\
  Tompson'15 \cite{conf/cvpr/TompsonGJLB15}  & 96.1  & 91.9  & 83.9  & 77.8  & 80.9  & 72.3  & 64.8 & 82.0\\
  H.\&R.'16 \cite{conf/cvpr/HuR16}  & 95.0  & 91.6  & 83.0  & 76.6  & 81.9  & 74.5  & 69.5 & 82.4\\
  Pishchulin'13 \cite{conf/iccv/PishchulinAGS13}  & 94.1  & 90.2  & 83.4  & 77.3  & 82.6  & 75.7  & 68.6 & 82.4\\
  Lifschitz'16~\cite{conf/eccv/LifshitzFU16}  & 97.8  & 93.3  & 85.7  & 80.4  & 85.3  & 76.6  & 70.2 & 85.0\\
  Gkioxari'16~\cite{conf/eccv/GkioxariTJ16}  & 96.2  & 93.1  & 86.7  & 82.1  & 85.2  & 81.4 &  74.1 & 86.1\\
  Rafi'16~\cite{BMVC2016}  & 97.2  & 93.9  & 86.4  & 81.3  & 86.8  & 80.6  & 73.4 & 86.3\\
  Insafutdinov'16 \cite{conf/eccv/InsafutdinovPAA16}  & 96.8 &  95.2 &  89.3  & 84.4  & 88.4  & 83.4 &  78.0 & 88.5\\
  Wei'16 \cite{conf/cvpr/WeiRKS16}  & 97.8  & 95.0  & 88.7  & 84.0  & 88.4 &  82.8  & 79.4 & 88.5\\
  B.\&T.'16 \cite{conf/eccv/BulatT16}  & 97.9  & 95.1  & 89.9 &  85.3  & 89.4 &  85.7  & 81.7 & 89.7\\
 Newell'16  \cite{conf/eccv/NewellYD16}  & 98.2  & 96.3  & 91.2  & 87.1  & 90.1  & 87.4  & 83.6 & 90.9\\
  Yang'17 \cite{chu2017multi}  & \textbf{98.5}  & 96.3  & 91.9  & 88.1  & \textbf{90.6}  & 88.0  & 85.0 & 91.5\\

Chou'17			\textcolor{black}{\cite{chou2017self}\footnotemark[4]}  & \textcolor{black}{{98.2}}& \textcolor{black}{96.8}& \textcolor{black}{92.2}& \textcolor{black}{88.0}& \textcolor{black}{91.3}& \textcolor{black}{89.1}& \textcolor{black}{84.9}&\textcolor{black}{91.8}\\
Yang'17      \textcolor{black}{\cite{yang2017learning}}\footnotemark[5]& \textcolor{black}{98.4}  & \textcolor{black}{96.5}  & \textcolor{black}{91.9}  & \textcolor{black}{88.2}  & \textcolor{black}{91.1}  & \textcolor{black}{88.6}  & \textcolor{black}{85.3} & \textcolor{black}{91.8}\\
Ke'18  \textcolor{black}{\cite{ke2018multi}}\footnotemark[6]& \textcolor{black}{98.5}  & \textcolor{black}{96.8}  & \textcolor{black}{92.7}  & \textcolor{black}{88.4}  & \textcolor{black}{90.6}  & \textcolor{black}{89.3}  & \textcolor{black}{86.3} & \textcolor{black}{92.1}\\

  \hline
  \hline
    Ours (test)\footnotemark[1]& {98.1}& \textbf{96.5}& \textbf{92.5}& \textbf{88.5}& {90.2}& \textbf{89.6}& \textbf{86.0}&\textbf{91.9}\\
    \hline
    Ours (-valid)\footnotemark[2]& {98.2}& {96.2}& {90.9}& {86.7}& {89.8}& {87.0}& {83.2}&{90.6}\\
    Ours (valid)\footnotemark[3]& {98.6}& {96.4}& {92.4}& {88.6}& {91.5}& {88.6}& {85.7}&{92.1}\\
  \hline
 \end{tabular}

\leftline{
\footnotemark[1]{\footnotesize{Our full model on test set.}}
\footnotemark[2]{\footnotesize{Our baseline model on validation set.}}}
\leftline{
\footnotemark[3]{\footnotesize{Our full model on validation set.}}}
 \leftline{
\footnotemark[4]{\footnotesize{The version using the same training set as our method and \cite{conf/cvpr/TompsonGJLB15}}.	}}
\leftline{
\footnotemark[4]$^,$\footnotemark[5]$^,$\footnotemark[6]{\footnotesize{Published after the submission of our conference version.}}}
\end{table}

\noindent \textbf{Experimental Settings.} According to the rough person location given by the dataset, we crop the images with the target human centered at the images, and warp the image patch to the size of 256$\times$256 pixels. We follow the data augmentation in~\cite{conf/eccv/NewellYD16} by rotation (+/- 30 degrees), and scaling (0.75-1.25). During training for LSP, we use the MPII dataset to augment the training data of LSP, which is a regular routine as done in~\cite{conf/cvpr/WeiRKS16,conf/eccv/InsafutdinovPAA16}.

During testing on the MPII dataset, we follow the standard routine to crop image patches with the given rough position and scale. The network starts with a 7$\times$7 convolutional layer with stride 2, followed by a residual modules and a max pooling to drop the resolution down from 256 to 64. Then two residual modules are followed before sending the feature into \textit{G}. Across the entire network all residual modules contain three convolution layers and a skip connection with output of 512 feature maps. The generator is stacked four times if not specially indicated in our experiment.

The network is trained using the RMSprop algorithm with initial learning rate of $2.5\times 10 ^{-4}$. The model on the MPII dataset was trained for 230 epochs and the LSP dataset for 250 epochs (about 2 and 3 days on a Tesla M40 GPU).

Next, we  perform  analysis on single-person human pose estimation and results for multi-person pose estimation are provided in the ablation study.

\subsubsection{Quantitative Results}\label{subsec:Results}

We use the Percentage Correct Keypoints (PCK@0.2) \cite{yang2013articulated} metric for comparison on the LSP dataset which reports the percentage of detection that falls within a normalized distance of the ground-truth for comparisons. For MPII, the distance is normalized by a fraction of the head size~\cite{conf/cvpr/AndrilukaPGS14} (referred to as PCKh@0.5). %

\noindent \textbf{LSP Human Pose.} Table~\ref{tab:Comparisons-of-PCK} shows the PCK performance of our method and some existing methods at a normalized distance of 0.2. Our method achieves the second best performance, and obtains 2.4\% improvement over previous methods in average.
In \cite{chou2017self} the authors also use adversarial training to improve the performance based on the
hourglass network. However, it uses the auto-encoder architecture for the discriminator and uses the reconstruction loss instead of classification loss compared to our method. Nevertheless,
it  shows the effectiveness of adversarial training for the task of pose estimation.

\noindent \textbf{MPII Human Pose.} Table~\ref{tab:Results-on-MPII} and Fig.~\ref{fig:PCKh-comparison-on} report
the PCKh performance of our method and previous methods at a normalized distance of 0.5. The
baseline model here refers to a four-stacked single-task network without multi-task and discriminators.
It has similar structure but half of stacked layers and parameter numbers compared to~\cite{conf/eccv/NewellYD16}. Our method achieves the best PCKh score of 91.9\% on the test set.

In particular, for the most challenging body parts, \emph{e.g.}, wrist and ankle, our method achieves 0.4\% and 1.0\% improvement compared with the closest competitor
respectively.

Note that, recently, after the submission of this manuscript,
the performance on this dataset has been further improved.
Ke \emph{et al.}~\cite{ke2018multi} also emphasize on the importance of structural learning and design an structural heatmap loss,  reporting a  PCKh score of 92.1\%  on the test  set. The work of \cite{yang2017learning} uses  a pyramid structure to enrich the DCNN features in scale and achieves similar PCKh scores with our method.

\begin{figure*}[!t]
\centering
\includegraphics[width=0.93\textwidth]{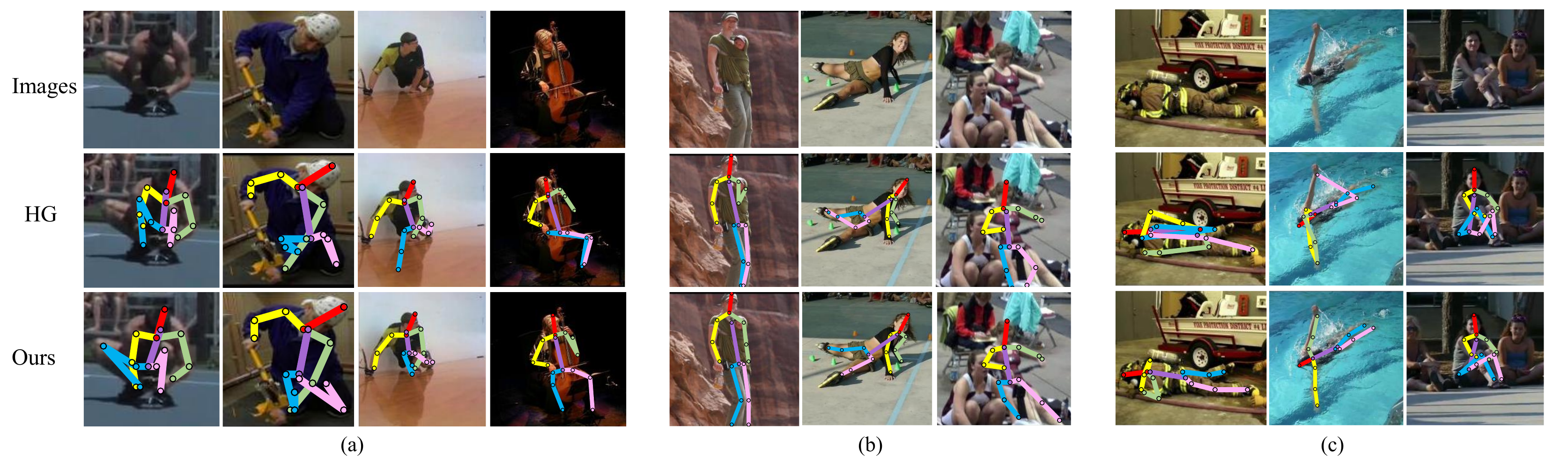}
\caption{Prediction samples on the MPII test set. First row: original images.
Second row: results by stacked hourglass network (HG)~\cite{conf/eccv/NewellYD16}.
Third row: results by our method. (a)-(c) show three different  failure cases  of HG.}
\label{fig:Prediction-samples-on}
\end{figure*}

\subsubsection{Qualitative Comparisons}\label{subsec:Comparison-of-Predicted}

To gain insights on how the proposed method accomplishes the goal of setting the pose estimations  within the geometric constraints, we visualize the predicted poses on the MPII test set compared with a 2-stacked hourglass network (HG)~\cite{conf/eccv/NewellYD16}, as demonstrated in Fig.~\ref{fig:Prediction-samples-on}. For fair comparison, we also use a 2-stacked network as baseline for this experiment.
We can see that our method indeed  takes the structure information of  the human body
into consideration, leading  to plausible  predictions.

In (a), the human body is highly twisted or partly occluded, which results in some invisible body limbs. In these cases, HG fails to understand some poses while our method succeeds. This may be because of the ability of occlusion prediction and shape prior
learned  in the training process. In (b), HG locates some body parts to the nearby positions with the most salient features. This indicates that HG has learned excellent features about body parts. However, without human body structure awareness,
it may locate some body parts to the surrounding area instead of the right one. In (c), due to the
lack of body configuration constraints, HG produces poses with strange twisting across body limbs.
As we have implicitly embedded the body constraints into our discriminator, our network manages to predict
the correct body location even under some difficult situations.

\begin{figure*}[t!]
\centering
\includegraphics[width=1.05\columnwidth]{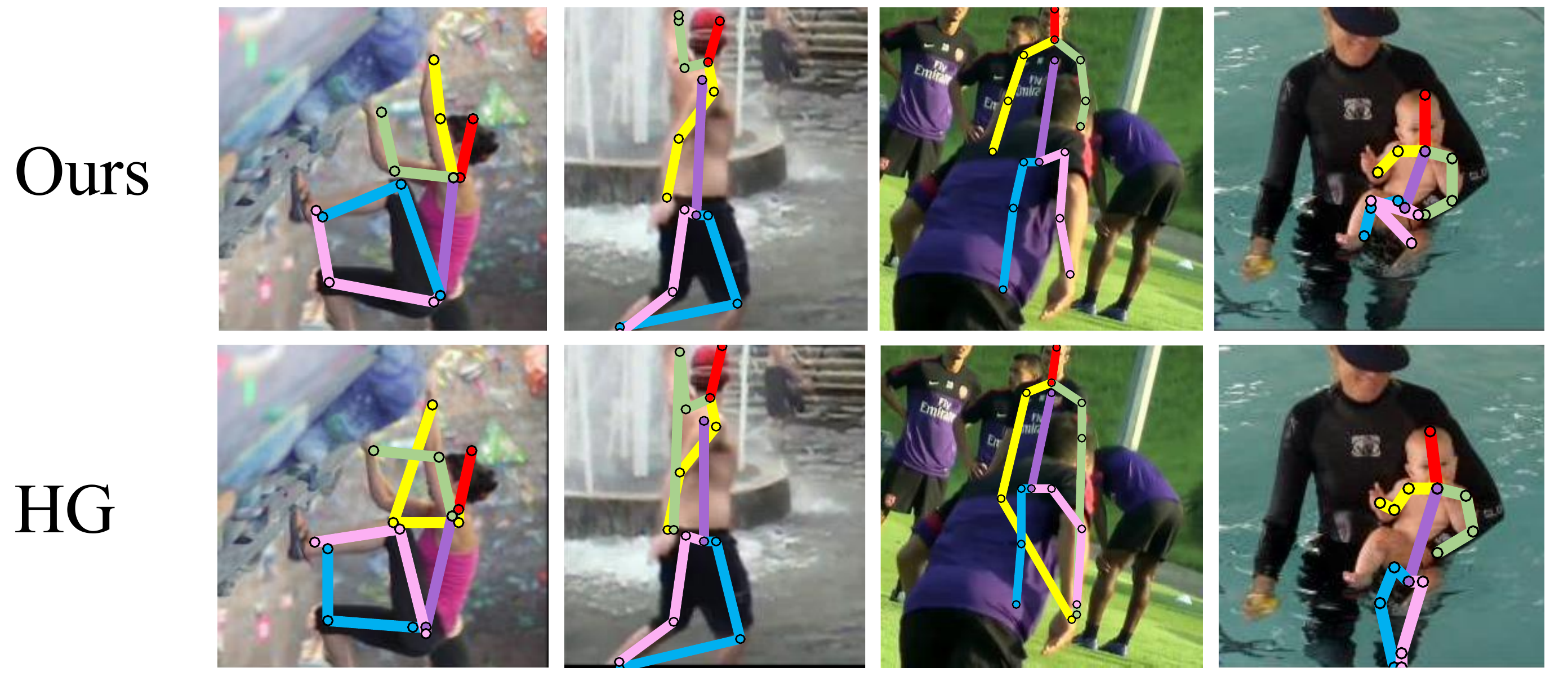}
\caption{Failure cases caused by body parts at the edge  (first and second columns), overlapping people (third column) and invisible limbs (the fourth column).
The results on the first and second rows are generated by our method and HG~\cite{conf/eccv/NewellYD16}, respectively.}
\label{fig:Failure-cases-caused}
\end{figure*}

We also show some failure examples of our method on the MPII test set in Fig.~\ref{fig:Failure-cases-caused}. As shown in Fig.~\ref{fig:Failure-cases-caused}, our method may fail in some challenging cases with twisted limbs at the edge, overlapping people and occluded body parts. In some cases, human may also fail to figure out the correct pose at a glance.
Even when our method fails in this situations, it still achieves more reasonable poses compared to previous methods.
Previous method may generate some poses which violate human body structure as shown in the first row of Fig.~\ref{fig:Failure-cases-caused}. When the network fails to find high-confidence locations around the person, it shifts to the surrounding area where the local features matches the trained features best. Lacking of shape constraint finally results in these strange  poses.


\begin{figure*}[h!]
\centering
\includegraphics[width=1.490\columnwidth]{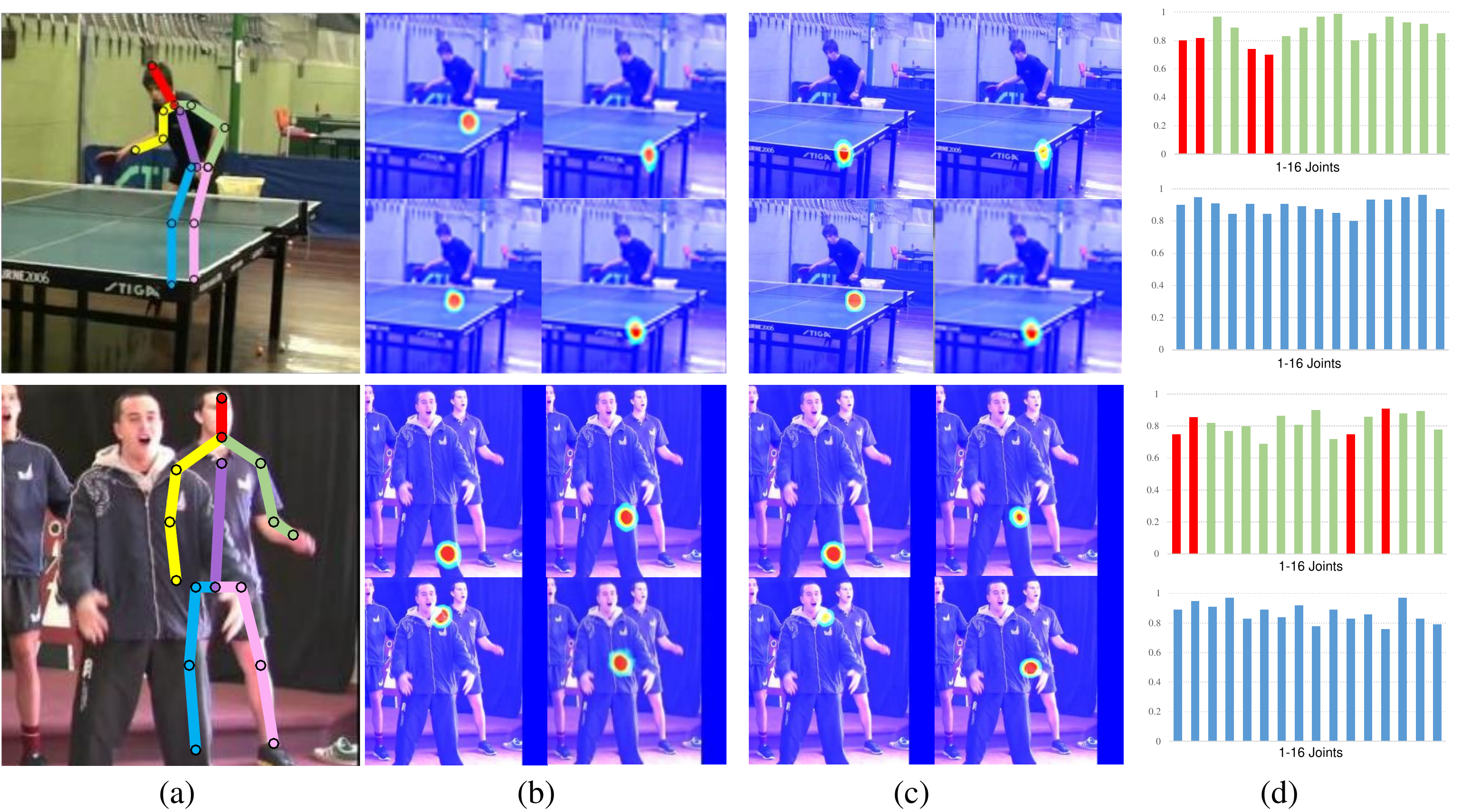}
\caption{(a) Input images with predicted poses; (b) Predicted pose heatmaps of four occluded body parts; (c) Predicted occlusion heatmaps of four occluded body parts; (d) Outputs values of \textit{P} (in blue) and \textit{C} (in green).
Red bars in the output of \textit{C} correspond to values of the four occluded body parts.}
\label{fig:Examples-inputs-and}
\end{figure*}

\subsubsection{Ablation Study}

To investigate the efficacy of the proposed multi-task generator network and the  discriminators designed for learning human body priors, we conduct ablation experiments on the validation set of the MPII Human Pose dataset and MSCOCO keypoint dataset. Analysis about occlusion, multi-task learning and discriminators are given as follows.

\begin{table}[t!]
 \caption{Detection rates of visible and invisible elbows and wrists.} \label{tab:Detection-rates-of}
 \small
 \centering
 \renewcommand\arraystretch{1}
 \begin{tabular}{ r ||cc|cc }
  \hline
  \multirow{2}{*}{Methods} & \multicolumn{2}{c}{Visible} & \multicolumn{2}{c}{Invisible}\\
\cline{2-5} & \emph{Wrist} & \emph{Elbow} & \emph{Wrist} & \emph{Elbow} \\
  \hline
  \hline
 ~\cite{conf/eccv/NewellYD16} & 93.6  &  95.1  &  67.2  &  74.0 \\
  \hline
  Ours &  \textbf{94.5}  &  \textbf{95.9}   & \textbf{70.7}  &  \textbf{77.6}  \\
  \hline
 \end{tabular}
\end{table}

\begin{table}[t]
 \caption{Results on the MSCOCO keypoint detection test-dev set.} \label{tab:MSCOCO2}
 \small
 \centering
 \renewcommand\arraystretch{1}
 \begin{tabular}{ r  |c|c|c|c|c }
  \hline
  \multirow{1}{*}{Methods} & {\textbf{AP}} & \multicolumn{1}{c}{AP$^{50}$}  & \multicolumn{1}{c}{AP$^{75}$} & \multicolumn{1}{c}{AP$^{M}$}  & \multicolumn{1}{c}{AP$^{L}$}\\
  \hline
CMU-Pose ~\cite{cao2017realtime} & 61.8 &  \multicolumn{1}{c}{84.9}  &  \multicolumn{1}{c}{67.5}  &  \multicolumn{1}{c}{57.1} &  \multicolumn{1}{c}{68.2}\\
G-RMI~\cite{papandreou2017towards} &  68.5  &  \multicolumn{1}{c}{87.1}  &  \multicolumn{1}{c}{75.5}  &  \multicolumn{1}{c}{65.8}  &  \multicolumn{1}{c}{73.3}\\
Mask R-CNN~\cite{he2017mask} &  63.1  &  \multicolumn{1}{c}{87.3}  &  \multicolumn{1}{c}{68.7}  &  \multicolumn{1}{c}{57.8}  &  \multicolumn{1}{c}{71.4}\\
Megvii~\cite{chen2017cascaded} &  72.1 &  \multicolumn{1}{c}{91.4}  &  \multicolumn{1}{c}{80.0}  &  \multicolumn{1}{c}{68.7}  &  \multicolumn{1}{c}{77.2}\\
RMPE~\cite{fang2017rmpe} &  61.8  &  \multicolumn{1}{c}{83.7}  &  \multicolumn{1}{c}{69.8}  &  \multicolumn{1}{c}{58.6}  &  \multicolumn{1}{c}{67.6}\\
RMPE++~\cite{fang2017rmpe} &  72.3  &  \multicolumn{1}{c}{89.2}  &  \multicolumn{1}{c}{79.1}  &  \multicolumn{1}{c}{68.0}  &  \multicolumn{1}{c}{78.6}\\
      \hline
Baseline &  68.4  &  \multicolumn{1}{c}{86.5}  &  \multicolumn{1}{c}{74.7}  &  \multicolumn{1}{c}{63.6}  &  \multicolumn{1}{c}{75.7}\\
Ours &  70.5  &  \multicolumn{1}{c}{88.0}  &  \multicolumn{1}{c}{76.9}  &  \multicolumn{1}{c}{66.0}  &  \multicolumn{1}{c}{77.0}\\

  \hline
 \end{tabular}
\end{table}

\begin{table*}[!t]
 \caption{Results on Human3.6M under Protocol \#1 (no rigid alignment in post-processing). SA indicates that a model was trained for each action, and MA indicates that a single model was trained for all actions.For 3d baseline and our method, SH indicates that the 2D poses are estimated using the Stacked Hourglass Network, GT indicates that the ground-truth 2D poses are used. As using ground-truth 2D pose is not fair for comparison with other methods, it is only used for evaluation of adversarial learning over 3D baseline.}

 \label{tab:Results-on-Human3.6M_no_rigid}
 \small
 \centering
 \setlength{\tabcolsep}{0.96pt}
 \renewcommand{\arraystretch}{1.25}
 \begin{tabular}{ r ||c|c|c|c|c|c|c|c|c|c|c|c|c|c|c||c }
  \hline
   {Methods} & \emph{Direct.} & \emph{Discuss} & \emph{Eating} & \emph{Greet} & \emph{Phone} & \emph{Photo} & \emph{Pose} & \emph{Purch.} & \emph{Sitting} & \emph{SitingD} &\emph{Smoke} &\emph{Wait} &\emph{WalkD} &\emph{Walk} &\emph{WalkT}  &\textbf{Mean} \\
  \hline
  \hline
  LinKDE\cite{ionescu2014human3}(SA) & 132.7 &  183.6  & 132.3  & 164.4  & 162.1  & 205.9  & 150.6 & 171.3 & 151.6 &243.0 & 162.1  & 170.7  & 177.1 & 96.6 & 127.9 &162.1\\
  Li \emph{et al.}~\cite{li2015maximum} (MA) & - &  136.9  & 96.9  & 124.7  & -  & 168.7  & - & - & - &- & -  & -  & 132.2 & 70.0 & - &-\\
  Tekin \emph{et al.}~\cite{tekin2016direct}(SA)  & 102.4 &  147.2  & 88.8  & 125.3  & 118.0  & 182.7  & 112.4 & 129.2 & 138.9 &224.9 & 118.4  & 138.8  & 126.3 & 55.1 & 65.8 &125.0\\
  Zhou\emph{et al.}~\cite{conf/cvpr/HuR16} (MA) & 87.4 &  109.3  & 87.1  & 103.2  & 116.2  & 143.3  & 106.9 & 99.8 & 124.5 &199.2 & 107.4  & 118.1  & 114.2 & 79.4 & 97.7 &113.0\\
  Tekin \emph{et al.}~\cite{tekin2016structured} (SA) & - &  129.1  & 91.4  & 121.7  & -  & 162.2  & - & - & - &- & -  & -  & 130.5 & 65.8 & - &-\\
  DeepViewPnt \cite{ghezelghieh2016learning} (SA) & 80.3 &  80.4  & 78.1  & 89.7  & -  & -  & - & - & - &- & -  & -  & - & 95.1 & 82.2 &-\\
  Du \emph{et al.}~\cite{du2016marker} (SA) & 85.1 &  112.7  & 104.9 & 122.1  & 139.1  & 135.9  & 105.9 & 166.2 & 117.5 &226.9 & 120.0  & 117.7  & 137.4 & 99.3 & 106.5 &126.5\\
  Park \emph{et al.}~\cite{park20163d} (SA) & 100.3 &  116.2  & 90.0  & 116.5  & 115.3 & 149.5  & 117.6 & 106.9 & 137.2 &190.8 & 105.8  & 125.1  & 131.9 & 62.6 & 96.2 &117.3\\
  Zhou \emph{et al.}~\cite{zhou2016deep} (MA)& 91.8 &  102.4  & 96.7  & 98.8  & 113.4  & 125.2  & 90.0 & 93.8 & 132.2 &159.0 & 107.0  & 94.4  & 126.0 & 79.0 & 99.0 &107.3\\
  Pavlakos \emph{et al.} (MA) \cite{pavlakos2016coarse}  & 67.4 &  71.9 & 66.7 & 69.1  & \textbf{72.0} & \textbf{77.0}  & 65.0 & 68.3 & 83.7 &\textbf{96.5} & 71.7  & 65.8  & 74.9& 59.1 & 63.2 &71.9\\
	 \textcolor{black}{Wei \emph{et al.}(MA) \cite{yang20183d}}\textcolor{black}{\footnotemark[1]}  & \textcolor{black}{51.5} &  \textcolor{black}{58.9} & \textcolor{black}{50.4} & \textcolor{black}{57.0}  & \textcolor{black}{62.1} & \textcolor{black}{65.4}  & \textcolor{black}{49.8} & \textcolor{black}{52.7} &  \textcolor{black}{69.2} &\textcolor{black}{ 85.2} & \textcolor{black}{57.4}  & \textcolor{black}{58.4}  &  \textcolor{black}{43.6}& \textcolor{black}{60.1} & \textcolor{black}{47.7} &\textcolor{black}{58.6}\\

  \hline
  \hline
  3d baseline (SH)(MA) \cite{martinez2017simple}  & 53.3 &  60.8  & 62.9  & 62.7  & 86.4  & 82.4 & 57.8 & 58.7 & 81.9 &99.8 & 69.1  & 63.9  & 67.1 & 50.9 & 54.8 &67.5\\
      Ours (SH)(MA) & \textbf{49.1} &  \textbf{58.8}  & \textbf{56.9 } & \textbf{60.2}  & 83.0  & 80.1 & \textbf{53.1} & \textbf{57.2} & \textbf{80.5} &\textbf{96.5} & \textbf{68.5}  & \textbf{61.9}  & \textbf{66.2} & \textbf{47.8} & \textbf{53.8} &\textbf{64.9}\\
   \hline
   3d baseline (GT)(MA) \cite{martinez2017simple}  & 37.7 &  44.4  & 40.3  & 42.1  & 48.2 & 54.9 & 44.4 & 42.1 & 54.6 &58.0 & 45.1  & 46.4 & 47.6 & 36.4 & 40.4 &45.5\\
     Ours (GT)(MA) & {36.2} & {43.8}  & {40.2 } & {40.2}  & 47.5  & 54.2 & {41.7} & {41.2} & {53.6} &{57.0} & {44.7}  & {45.1}  & {46.1} & {36.1} & {40.0} &{44.5}\\

  \hline
 \end{tabular}
\footnotemark[1]{\footnotesize{Published after the submission this manuscript.}}
\end{table*}

\begin{table*}[!t]
 \caption{Results on Human3.6M under Protocol \#2 (rigid alignment in post-processing). SA indicates that a model was trained for each action, and MA indicates that a single model was trained for all actions. SH indicates that the 2D poses are estimated using the Stacked Hourglass Network.}
 \label{tab:Results-on-Human3.6M_rigid}
 \small
 \centering
 \setlength{\tabcolsep}{0.6pt}
 \renewcommand{\arraystretch}{1.25}
 \begin{tabular}{ r  ||c|c|c|c|c|c|c|c|c|c|c|c|c|c|c||c }
  \hline
   {Methods} & \emph{Direct.} & \emph{Discuss} & \emph{Eating} & \emph{Greet} & \emph{Phone} & \emph{Photo} & \emph{Pose} & \emph{Purch.} & \emph{Sitting} & \emph{SitingD} &\emph{Smoke} &\emph{Wait} &\emph{WalkD} &\emph{Walk} &\emph{WalkT}  &\textbf{Mean} \\
  \hline
  \hline
  Akhter \& Black\cite{akhter2015pose} (SA) & 199.2 &  177.6  & 161.8  & 197.8  & 176.2  & 186.5  & 195.4 & 167.3 & 160.7  & 173.7  & 177.8 & 181.9 & 176.2 & 198.6 &192.7 &181.1\\
  Ramakrishna \emph{et al.} \cite{ramakrishna2012reconstructing} (MA) & 137.4 &  149.3  & 141.6  & 154.3 & 157.7  & 158.9  & 141.8 & 158.1 & 168.6 &175.6 & 160.4  & 161.7  & 150.0 & 174.8 & 150.2&157.3\\
  Zhou \emph{et al.} \cite{zhou2017sparse} (SA)  & 99.7 &  95.8  & 87.9  & 116.8  & 108.3  & 107.3  & 93.5 & 95.3 & 109.1 &137.5 & 106.0  & 102.2  & 106.5& 110.4 & 115.2 &106.7\\
  Bogo\cite{bogo2016keep} (MA) & 62.0&  60.2  & 67.8  & 76.5  & 92.1  & 77.0 & 73.0 & 75.3 & 100.3 &137.3 & 83.4  & 77.3  & 86.8 & 79.7 & 87.7 &82.3\\
  Moreno-Noguer \cite{moreno20163d} (SA) & 66.1 &  61.7  & 84.5  & 73.7  & 65.2 & 67.2 & 60.9& 67.3& 103.5 &74.6 & 92.6  & 69.6  & 71.5 & 78.0 & 73.2 &74.0\\
  Pavlakos{et al.} \cite{pavlakos2016coarse} (SA) & -&  -  & -  & -  & -  & -  & - & - & - &- & -  & -  & - & -& - &51.9\\
  3d baseline (SH)(MA) \cite{martinez2017simple}  & 39.5&  43.2  & 46.4 & 47.0 & 51.0  & 56.0 & 41.4 & 40.6 & 56.5 &69.4 & 49.2  & 45.0 & 49.5& 38.0 & 43.1 &47.7\\
   Wei \emph{et al.} (MA)\cite{yang20183d}\footnotemark[1]  & 26.9 &  {30.9} & {36.3} & {39.9}  & {43.9} & {47.4}  & {28.8} &{29.4} &  {36.9} &{ 58.4} & {41.5}  &{30.5}  &  {29.5}& {42.5} & {32.2} &{37.7}\\
  \hline
  \hline
    Ours (SH)(MA) & \textbf{38.5} &  \textbf{42.7}  & \textbf{43.9 } & \textbf{46.1}  & \textbf{49.1}  & \textbf{53.2} & \textbf{41.0} & \textbf{39.8} & \textbf{53.9} &\textbf{63.8} & \textbf{48.1}  & \textbf{43.9}  & \textbf{49.3} & \textbf{37.6} & \textbf{41.0} &\textbf{46.1}\\
    \hline

 \end{tabular}
  \textcolor{black}{\footnotemark[1]{\footnotesize{Published after the submission of this manuscript.}}}

\end{table*}

\noindent\textbf{Occlusion Analysis}
Here we present  a detailed analysis of the outputs of the networks when joints in the images are occluded.
First, two examples with some body parts occluded are given in Fig.~\ref{fig:Examples-inputs-and}. In the first sample, two legs of the person are totally occluded by the table. In the corresponding occlusion maps, the occluded part are well predicted.
Despite  the occlusions, the pose heatmaps generated
by our method are mostly clear and Gaussian centered. This results in high scores in both pose prediction and confidence evaluation.

For the second image, half part of the person is overlapped by the person ahead of him. Our method again manages to predict the correct pose locations with clear heatmaps. Occlusion information is also well predicted for the occluded parts. As shown with the bars in red, although the confidence scores of the occluded body parts are relatively
low, they remain an overall high level. This shows that our network has learned  some degree of human body priors  during training. Thus it has the ability to predict plausible  poses even under some occlusions. This verifies our motivation of designing the discriminators with GANs.

Next, we compare the performance of our method under occlusions with a stacked hourglass network~\cite{conf/eccv/NewellYD16} as the strong baseline. In the validation set of MPII, about 25\% of the elbows and wrists with annotations are labeled invisible. We show the results of elbows and wrists with visible samples and invisible samples in Table~\ref{tab:Detection-rates-of}. For body parts without occlusions, our method improves the baseline by about 0.8\% of detection rate. However, {\it our method improves the baseline by 3.5\% and 3.6\% of detection rates on the invisible wrists and elbows. This shows the advantage of our method in dealing with body parts with occlusions.}

\begin{figure}[t!]
\centering
\includegraphics[width=0.85\columnwidth]{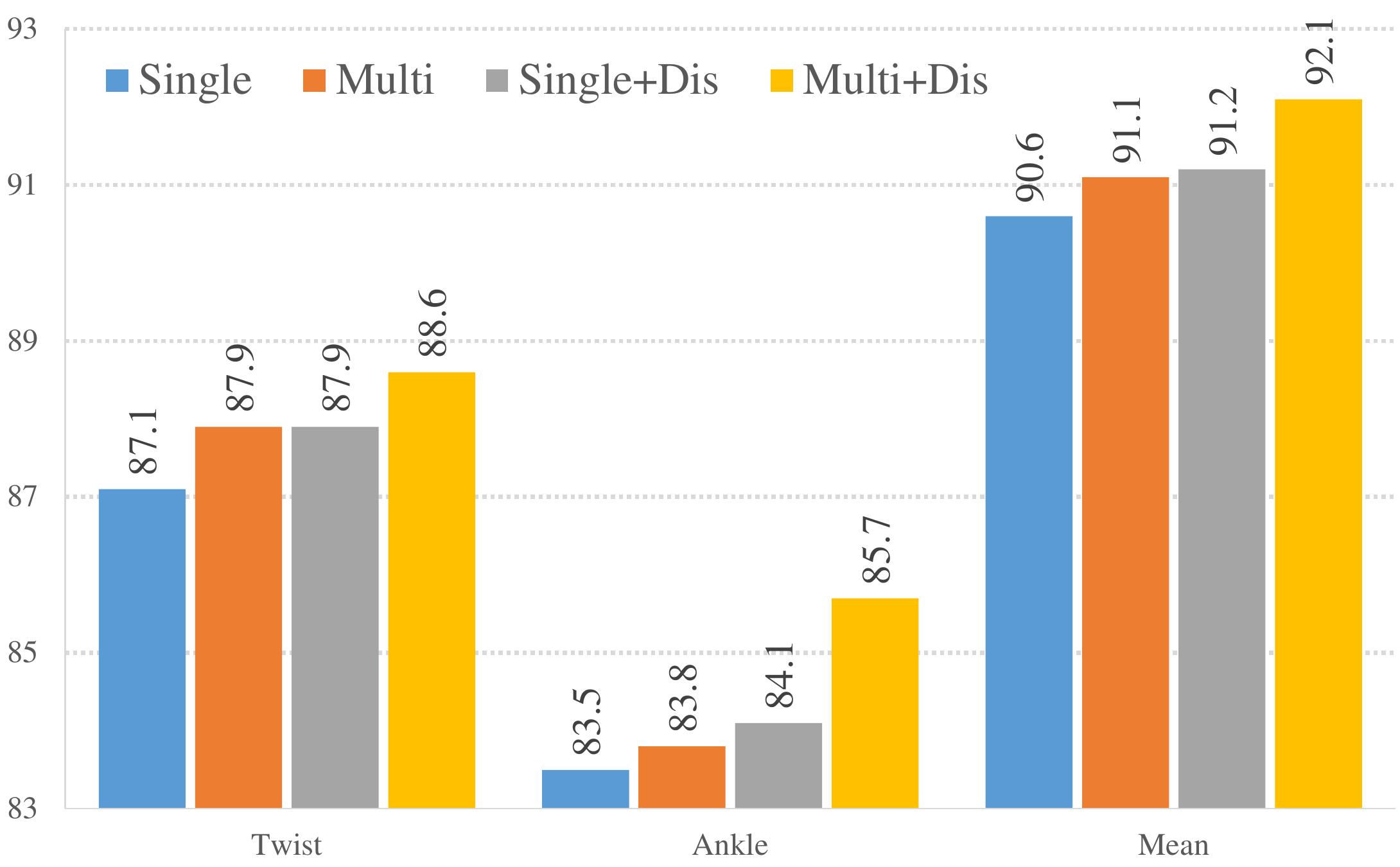}
\caption{Ablation study: PCKh scores at the threshold of 0.5.}
\label{fig:9}
\end{figure}

\noindent\textbf{Multi-task.} We compare the four-stacked multi-task generator with the single-task model. The networks are trained by removing the discriminators (\emph{i.e.}, no GANs). By using the occlusion information, the performance on the MPII validation set increases 0.5\% compared to the single-task model as shown in Fig.~\ref{fig:9}. This shows that the multi-task structure helps the network to understand the poses.

\begin{figure*}[t!]
\centering
\includegraphics[width=2\columnwidth]{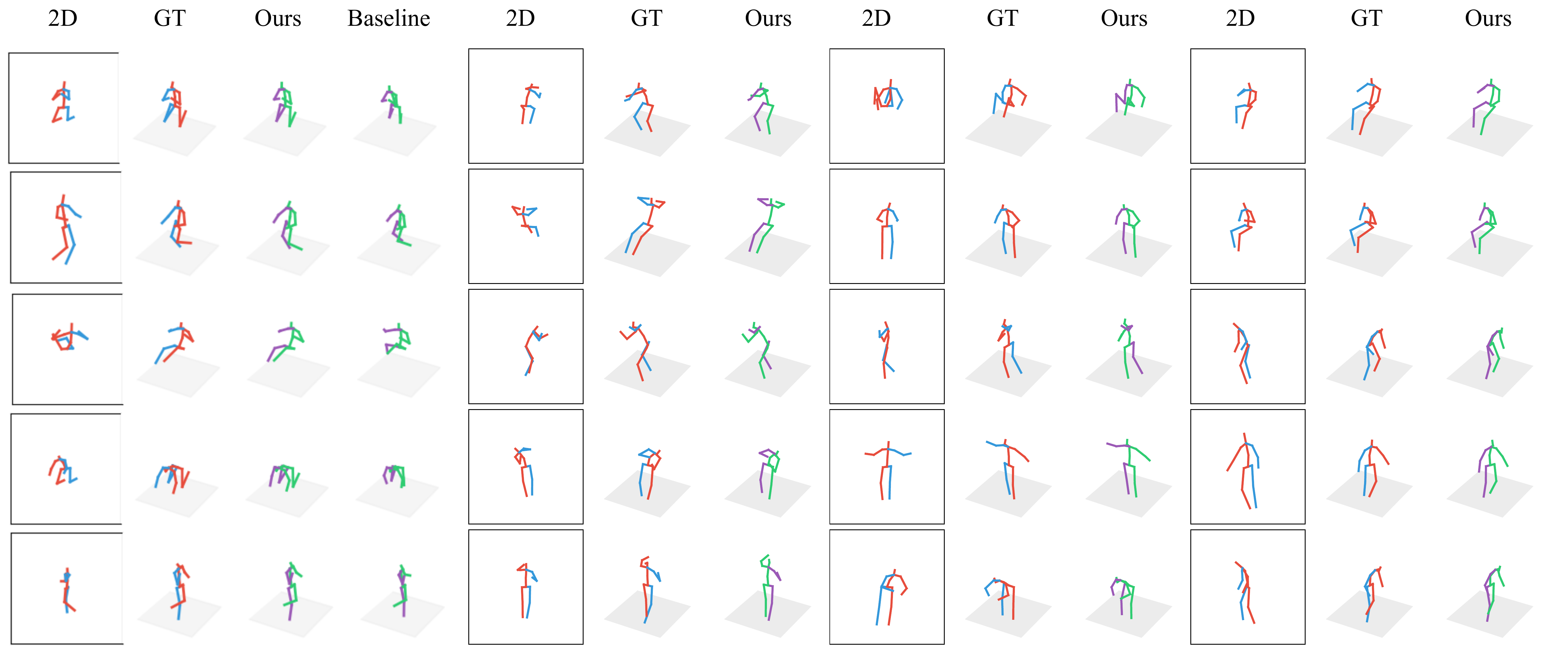}
\caption{3D pose examples on the Human 3.6M validation set. The left four columns show comparisons with baseline model. The proposed adversarial learning method refines the implausible poses generated by the baseline model and produces results more similar to ground-truth poses (GT). Other columns show 3D poses generated by our method in different scenes.}
\label{fig:Fig_3d_comparison}
\end{figure*}

\noindent\textbf{Discriminators.}
We first compare the four-stacked single-task generator trained with discriminators with the baseline.
The networks are trained by removing the part for the occlusion heatmaps.
Discriminators also receive inputs without occlusion heatmaps.
By using the body-structure-aware GANs, the performance on the MPII validation set increases by 0.6\% compared to the single-task model as in Fig.~\ref{fig:9}.

This shows that the discriminators  contribute in pushing the generator to produce more reliable pose predictions.
In general, individually adding the multi-task or discriminator both increases the accuracy of location.
But using them separately results in 0.6\% and 0.5\% improvement respectively, while using both produces an improvement of 1.5\%.
Occlusion information can clearly help  understand the image and generate  more accurate poses.

Second, we conducted experiments on the MPII validation set to show individual effects of \textit{P} and {\it C}. This is done by simply removing \textit{P} and \textit{C} separately in our method. With single \textit{P}, the performance of our method  evaluated by PCKh@0.5  is 91.9\% compared to 91.1\% by the baseline. With single \textit{C}, the performance is 91.4\%. It is clear that \textit{P} contributes more to our final improvement. \textit{P} incorporates information of whether the pose configuration is plausible. \textit{C} uses the same $\ell_2$ loss as baseline while using an adversarial learning strategy.

\noindent\textbf{Multi-person}
In the problem of multi-person pose estimation, occlusions and overlappings are more serious than single-person pose estimation.
Although our method focuses on single-person human pose estimation, we conduct ablation experiment on the MSCOCO  Keypoints dataset ~\cite{lin2014microsoft} to validate the effectiveness of our method under these circumstances.
The results are displayed in Table~\ref{tab:MSCOCO2}.
The performance of other methods is quoted from \cite{fang2017rmpe}.
The adversarial learning strategy  improves the performance of the baseline model.

\subsection{2D to 3D Pose Transformation}\label{subsec:3d pose}

\noindent \textbf{Datasets and Experimental Settings}
We focus our numerical evaluation on a public datasets for 3d human pose estimation: Human 3.6M ~\cite{ionescu2014human3}.
Human 3.6M is currently the largest publicly available datasets for human 3d pose estimation.
The dataset consists of 3.6 million images featuring 7 professional actors performing 15 everyday activities such as walking, eating, sitting, making a phone call and engaging in a discussion.
2d joint locations and 3d ground truth positions are available, as well as projection (camera) parameters and body proportions for all the actors.
We follow the standard protocol, using subjects 1, 5, 6, 7, and 8 for training, and subjects 9 and 11 for validation. For fair comparison to previous methods, we build our method based on a recently published baseline ~\cite{martinez2017simple} and strictly follow their experimental settings.

In detail, the 2D to 3D transformation net is the same as ~\cite{martinez2017simple}. 2D poses are estimated by the same  hourglass networks as ~\cite{martinez2017simple}used.
We only add our discriminators and adversarial training to provide structural information to the original method.
The average error in millimeters between the ground truth and our prediction across all joints and cameras are reported, after alignment of the root (central hip) joint.
In some of the baselines, the prediction has been further aligned with the ground truth via a rigid transformation (\emph{e.g.}, ~\cite{bogo2016keep,moreno20163d}).
We refer the experiment without further alignment as Protocol \#1 while the one with alignment as Protocol \#2. On the other hand, some recent methods have trained one model for all the actions, as opposed to building action-specific models instead of independent training and testing in each action.
We also show their results under these two circumstances.

Table ~\ref{tab:Results-on-Human3.6M_no_rigid} reports the results without further alignment and Table ~\ref{tab:Results-on-Human3.6M_rigid} reports the results with further alignment.
By simply adding adversarial learning on ~\cite{martinez2017simple}, the performance is improved.
The work in \cite{yang20183d} followed the same adversarial learning routine with our method and proposed more complex discriminators to tell whether its prediction is plausible. It demonstrates that \textit{the performance of adversarial learning framework can be further improved by designing better discriminators}.
It should be pointed out all the gain in performance comes with no additional computation cost during test.

We also show examples on Human 3.6M of both the baseline model and the proposed method to show the effectiveness of adversarial learning in Fig.~\ref{fig:Fig_3d_comparison}.
Some implausible 3D poses generated by the baseline model is well refined by our method.

\section{Conclusions}

In this paper, we have proposed a novel conditional adversarial network for pose estimation, which trains a pose generator with discriminator networks. The discriminators function as an expert who distinguishes plausible poses from unreasonable ones. By training the pose generator to deceive the expert that the generated pose is real, our network is more robust to occlusions, overlapping and twisting of pose components. In contrast to previous work using DCNNs for  pose estimation, our network is able to alleviate  the risk of localizing   human body parts onto the  matched features  without consideration of human body priors.

Although we need to train three sub-networks (\textit{G}, \textit{P}, \textit{C}) during training,
we only need to use \textit{G} net during testing. With a negligible  computation overhead, we achieve considerably better results on a few popular benchmark datasets.
We have also verified that our network can produce poses which are mostly within the manifold of human body shapes.

The method developed here can be immediately applied to other shape estimation problems using DCNNs with minimal modification.
The inputs of the discriminators can also be further improved to boost the discrimination ability.
More significantly, we believe that the use of conditional GANs as a tool to predict structured  output  or enforce
output dependency can be further developed to much more general structured output learning.

\section*{Acknowledgements}

This work was partially supported by the National Science Fund of China under Grant  U1713208, Program for Changjiang Scholars
and “111” Program AH92005.
This work was in part supported by an ARC Future Fellowship to C. Shen and an ARC DECRA Fellowship to L. Liu;
and ARC Grant CE140100016.

\begin{IEEEbiography}[{\includegraphics[width=1in,height=1.25in,clip,keepaspectratio]{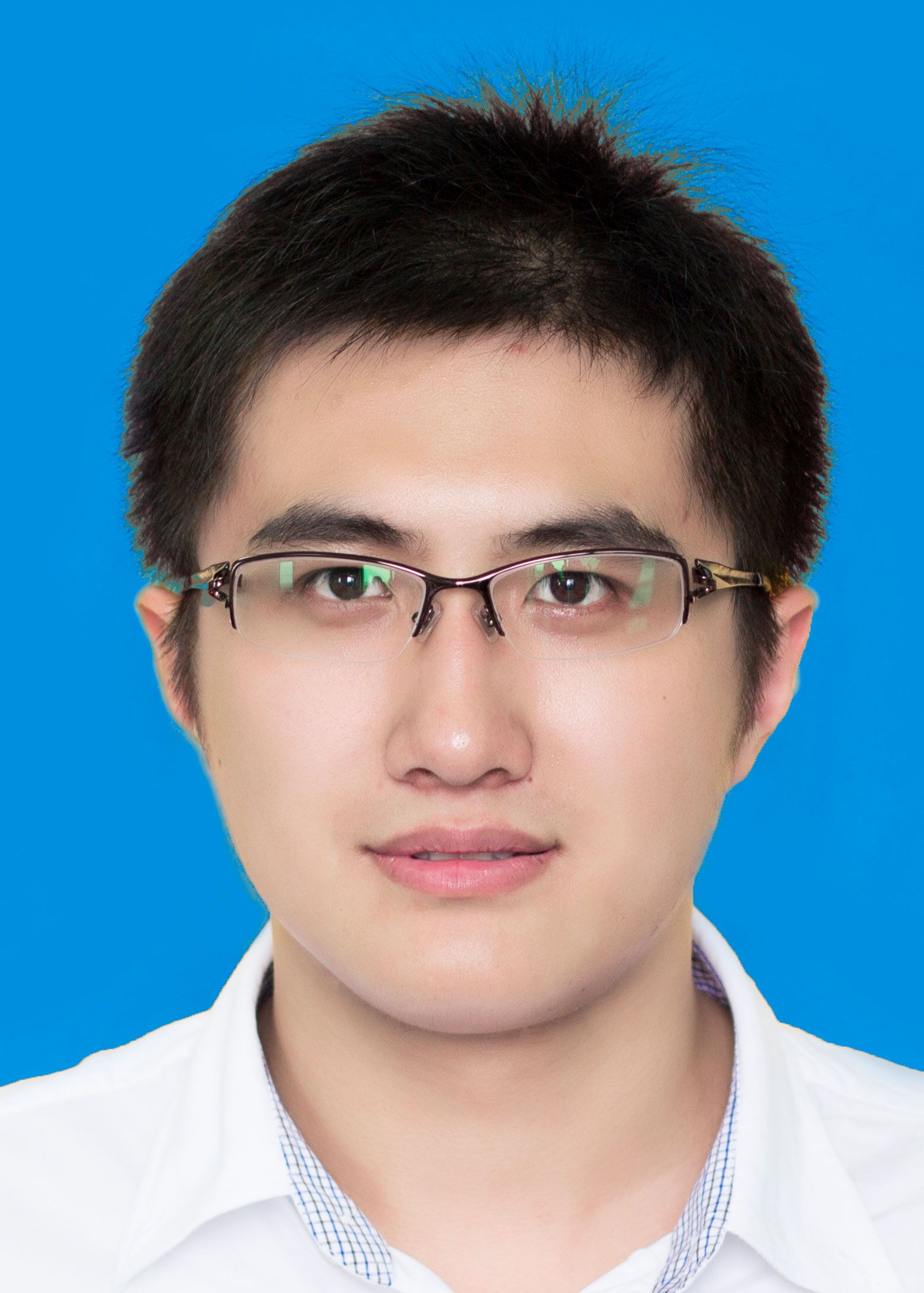}}]{Yu Chen}
received the BS degree in mathematics and applied mathematics from Nanjing University of Science and Technology.
He completed  the PhD degree at the same university.
He is now  a researcher at  Motovis Research Australia.
His current research interests are deep learning, autonomous driving and pose estimation in particular. \end{IEEEbiography}

\begin{IEEEbiography}
[{\includegraphics[width=1in,height=1.25in,clip,keepaspectratio]{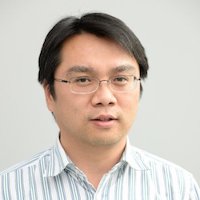}}]
{Chunhua Shen}
is a Professor at School of Computer Science, University of Adelaide.
Before that, he was with the computer vision program at NICTA (National ICT Australia), Canberra Research Laboratory for about six years.
He studied at Nanjing University, at Australian National University, and received his PhD degree from the University of Adelaide.
    From 2012 to 2016, he held an Australian Research Council Future Fellowship.
 \end{IEEEbiography}

\begin{IEEEbiography}
[{\includegraphics[width=1in,height=1.25in,clip,keepaspectratio]{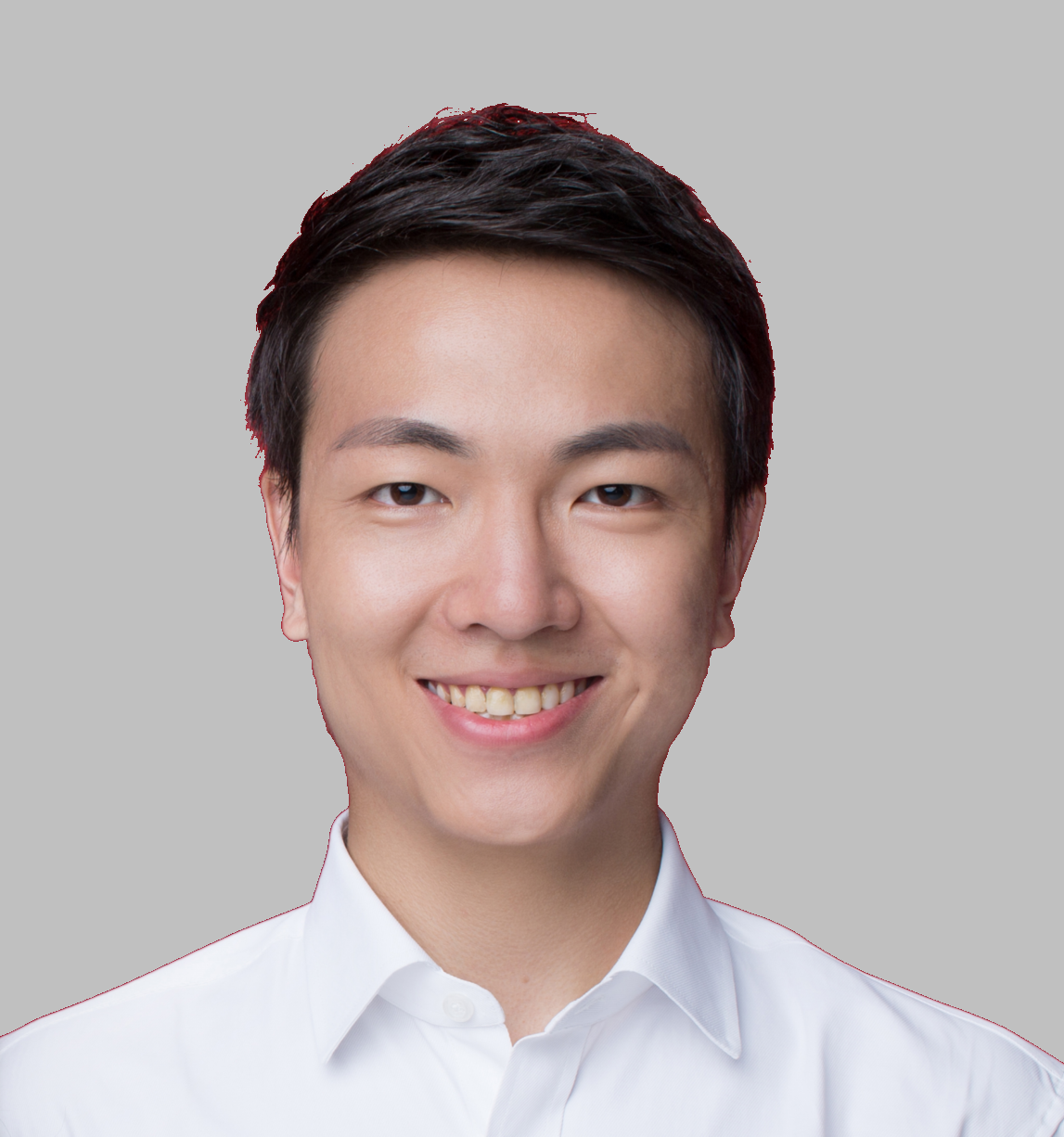}}]{Hao Chen} received the master's degree from Zhejiang University, China. He is working towards the PhD degree at School of Computer Science,  The University of Adelaide. His current research interests in deep learning and
 its applications in computer vision and text analysis.
 \end{IEEEbiography}

\begin{IEEEbiography}[{\includegraphics[width=1in,height=1.25in,clip,keepaspectratio]{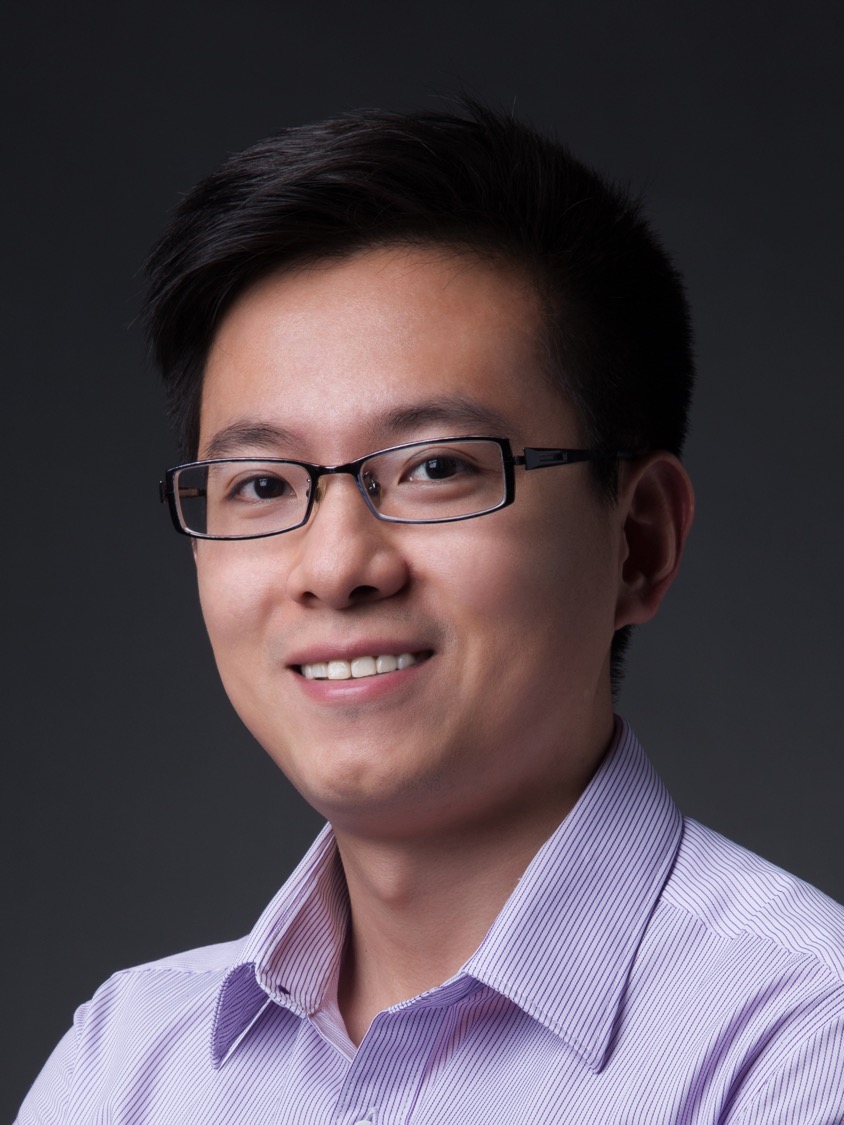}}]{Xiu-Shen Wei} (M'18) received his BS degree in computer science, and his Ph.D. degree in computer science and technology from Nanjing University. He is now the Research Lead of Megvii (Face++) Research Nanjing. He has published more than ten academic papers on the top-tier international journals and conferences, such as IEEE TIP, IEEE TNNLS, Machine Learning Journal, ICCV, IJCAI, etc. He achieved the first place in the Apparent Personality Analysis competition (in association with ECCV 2016) and the first runner-up in the Cultural Event Recognition competition (in association with ICCV 2015) as the team director. He also received the Presidential Special Scholarship (the highest honor for Ph.D. students) in Nanjing University. His research interests are computer vision and machine learning. He is a PC member of ICCV, CVPR, ECCV, NIPS, IJCAI, AAAI, etc..\end{IEEEbiography}

\begin{IEEEbiography}
[{\includegraphics[width=1in,height=1.25in,keepaspectratio]{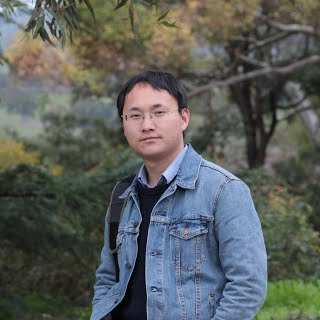}}]
{Lingqiao Liu} received the BS and MS degrees in communication engineering from the University of Electronic Science and Technology of China, Chengdu, in 2006 and 2009, respectively, and the PhD degree from the Australian National University, Canberra, in 2014. He is now a Lecturer at the University of Adelaide. In 2016, he was awarded the Discovery Early Career Researcher Award by the Australian Research Council. His research interests include various topics in computer vision and machine learning.
\end{IEEEbiography}

\begin{IEEEbiography}
[{\includegraphics[width=1in,height=1.25in,clip,keepaspectratio]{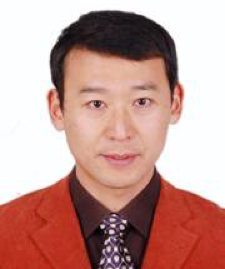}}]{Jian Yang} received the PhD degree from Nanjing University of Science and Technology (NUST), on the subject of pattern recognition and intelligence systems in 2002. In 2003, he was a postdoctoral researcher at the University of Zaragoza. From 2004 to 2006, he was a Postdoctoral Fellow at Biometrics Centre of Hong Kong Polytechnic University. From 2006 to 2007, he was a Postdoctoral Fellow at Department of Computer Science of New Jersey Institute of Technology. Now, he is a Chang-Jiang professor in the School of Computer Science and Technology of NUST. He is the author of more than 100 scientific papers in pattern recognition and computer vision. His journal papers have been cited more than 4000 times in the ISI Web of Science, and 9000 times in the Web of Scholar Google. His research interests include pattern recognition, computer vision and machine learning. Currently, he is/was an associate editor of Pattern Recognition Letters, IEEE Trans. Neural Networks and Learning Systems, and Neurocomputing. He is a Fellow of IAPR.\end{IEEEbiography}

\ifCLASSOPTIONcaptionsoff
  \newpage
\fi

\bibliographystyle{IEEEtran}
\bibliography{IEEEabrv,CSRef0}

\end{document}